\documentclass[10pt, twocolumn, a4paper, logo]{googledeepmind}

\usepackage[numbers, sort&compress, square]{natbib}
\usepackage{soul}
\makeatletter
\renewcommand{\bibfont}{\normalfont\small}
\makeatother

\usepackage[capitalise]{cleveref}
\creflabelformat{equation}{#2\textup{#1}#3}

\usepackage{fancybox}
\usepackage{makecell}
\usepackage{appendix}
\usepackage{hyperref}
\usepackage{comment}

\usepackage{amsfonts}
\usepackage{titlesec}
\usepackage{amsmath}
\usepackage{graphicx}
\usepackage{listings}
\usepackage{xcolor}
\usepackage{todonotes}
\usepackage{paralist}
\usepackage{authblk}
\usepackage{pifont}
\usepackage{ifthen}
\usepackage{float}
\usepackage{subcaption}
\usepackage{booktabs}
\usepackage{dblfloatfix}

\PassOptionsToPackage{inline}{enumitem}
\newlist{inlinelist}{enumerate*}{1}
\setlist[inlinelist]{label=(\textit{\roman*})}
\correspondingauthor{seijink@google.com, yaschimpf@google.com, schlegelm@google.com}

\author[*,1]{Seijin Kobayashi}
\author[*,1]{Yanick Schimpf}
\author[*,1]{Maximilian Schlegel}
\author[1]{\authorcr\vspace{3pt} Angelika Steger}
\author[1]{Maciej Wolczyk}
\author[1]{Johannes von Oswald}
\author[1]{Nino Scherrer}
\author[1]{Kaitlin Maile}
\author[1]{Guillaume Lajoie}
\author[1]{Blake A. Richards}
\author[1]{Rif A. Saurous}
\author[2]{James Manyika}
\author[1]{Blaise Agüera y Arcas}
\author[*,1]{Alexander Meulemans}
\author[*,1]{João Sacramento}

\affil[1]{Google, Paradigms of Intelligence Team}
\affil[2]{Google}

\affil[*]{Core contributor.}
\usepackage{thm-restate, amsmath, amssymb, amsfonts, mathtools}
\usepackage[ruled,vlined]{algorithm2e}

\usepackage{comment}

\DeclareMathSymbol{\smin}{\mathbin}{AMSa}{"39}

\def\eqref#1{equation~\ref{#1}}

\def\1{\bm{1}}

\DeclareMathAlphabet{\mathsfit}{\encodingdefault}{\sfdefault}{m}{sl}
\SetMathAlphabet{\mathsfit}{bold}{\encodingdefault}{\sfdefault}{bx}{n}

\usepackage{booktabs}
\usepackage{wrapfig}

\newcommand{\HparamsExpertAnt}{
\begin{table}[htb]
    \centering
    \footnotesize
    \begin{tabularx}{\columnwidth}{X l}
        \toprule
        \textbf{Hyperparameter} & \textbf{Search} \\
        \midrule
        Embedding dimension $n_e$ & 256 \\
        Backbone model & SSM \\
        Backbone model depth & 6 \\
        Train steps & 256000 \\
        Batch size & 1024 \\
        Optimizer & AdamW \\
        \hspace{3mm} Learning rate & [3e-4] \\
        \hspace{3mm} Weight decay & [0.03] \\
        \hspace{3mm} Entropy reg & [0.0003] \\
        \hspace{3mm} $\beta$s & (0.9, 0.999) \\
        Scheduler \\
        \hspace{3mm} Constant learning rate \\
        \bottomrule
    \end{tabularx}
    \caption{\textbf{Hyperparameters for expert training on the ant-pinpad environment.}}
    \label{tab:HparamsExpertAnt}
\end{table}
}

\newcommand{\HparamsBeliefGrid}{
\begin{table}[htb]
    \centering
    \footnotesize
    \begin{tabularx}{\columnwidth}{X l} 
        \toprule
        \textbf{Hyperparameter} & \textbf{Search} \\
        \midrule
        Embedding dimension $n_e$ & 256 \\
        Backbone train steps & [16K, 32K, 64K, 128K, 256K] \\
        Probe layers & [0, 1, 2, 3, 4, 5, 6] \\
        Train steps probe & 8000 \\
        Batch size & 512 \\
        Optimizer & AdamW \\
        \hspace{3mm} Learning rate & [1e-3] \\
        \hspace{3mm} Weight decay & [0.0] \\
        \hspace{3mm} $\beta$s & (0.9, 0.999) \\
        Scheduler \\
        \hspace{3mm} Constant learning rate \\
        \bottomrule
    \end{tabularx}
    \caption{\textbf{Hyperparameters for belief state probing in gridworld.}}
    \label{tab:HparamsBeliefGrid}
\end{table}
}

\newcommand{\HparamsBackboneGrid}{
\begin{table}[htb]
    \centering
    \footnotesize
    \begin{tabularx}{\columnwidth}{X l}
        \toprule
        \textbf{Hyperparameter} & \textbf{Search} \\
        \midrule
        Embedding dimension $n_e$ & 256 \\
        Backbone model & Transformer \\
        Backbone model depth & 6 \\
        Observation coefficient $\lambda$ & [0, 0.01] \\
        Train steps & 256000 \\
        Batch size & 1024 \\
        Optimizer & AdamW \\
        \hspace{3mm} Learning rate & [3e-4] \\
        \hspace{3mm} Weight decay & [0.03] \\
        \hspace{3mm} $\beta$s & (0.9, 0.999) \\
        Scheduler \\
        \hspace{3mm} Constant learning rate \\
        \bottomrule
    \end{tabularx}
    \caption{\textbf{Hyperparameters for sequence model training on the gridworld environment.}}
    \label{tab:HparamsBackboneGrid}
\end{table}
}

\newcommand{\HparamsBackboneAnt}{
\begin{table}[htb]
    \centering
    \footnotesize
    \begin{tabularx}{\columnwidth}{X l}
        \toprule
        \textbf{Hyperparameter} & \textbf{Search} \\
        \midrule
        Embedding dimension $n_e$ & 256 \\
        Backbone model & SSM \\
        Backbone model depth & 8 \\
        Observation coefficient $\lambda$ & 10 \\
        Train steps & 204800 \\
        Batch size & 512 \\
        Optimizer & AdamW \\
        \hspace{3mm} Learning rate & 3e-4 \\
        \hspace{3mm} Weight decay & 0.03 \\
        \hspace{3mm} $\beta$s & (0.9, 0.999) \\
        Scheduler \\
        \hspace{3mm} Constant learning rate \\
        \bottomrule
    \end{tabularx}
    \caption{\textbf{Hyperparameters for sequence model training on the ant-pinpad environment.}}
    \label{tab:HparamsBackboneAnt}
\end{table}
}

\newcommand{\HparamsControllerCompGrid}{
\begin{table}[htb]
    \centering
    \footnotesize
    \begin{tabularx}{\columnwidth}{X l}
        \toprule
        \textbf{Hyperparameter} & \textbf{Search} \\
        \midrule
        Observation coefficient $\lambda$ & 0 \\
        Controller model & Low-rank ($16$) linear\\
        Controlled layer $l$ & $\frac{L}{2}$ \\
        Train steps & 3200 \\
        Batch size & 512 \\
        Optimizer & AdamW \\
        \hspace{3mm} Learning rate & [1e-3] \\
        \hspace{3mm} Weight decay & [0.03] \\
        \hspace{3mm} $\beta$s & (0.9, 0.999) \\
        Scheduler \\
        \hspace{3mm} Constant learning rate \\
        \bottomrule
    \end{tabularx}
    \caption{\textbf{Hyperparameters for the controller compositional generalization experiment on the gridworld environment.}}
    \label{tab:HparamsControllerCompGrid}
\end{table}
}

\newcommand{\HparamsControllerCompAnt}{
\begin{table}[htb]
    \centering
    \footnotesize
    \begin{tabularx}{\columnwidth}{X l}
        \toprule
        \textbf{Hyperparameter} & \textbf{Search} \\
        \midrule
        Observation coefficient $\lambda$ & 0 \\
        Controller model & Linear\\
        Controlled layer $l$ & $\frac{L}{2}$ \\
        Train steps & 3200 \\
        Batch size & 256 \\
        Optimizer & AdamW \\
        \hspace{3mm} Learning rate & [3e-4] \\
        \hspace{3mm} Weight decay & [0.03] \\
        \hspace{3mm} $\beta$s & (0.9, 0.999) \\
        Scheduler \\
        \hspace{3mm} Constant learning rate \\
        \bottomrule
    \end{tabularx}
    \caption{\textbf{Hyperparameters for the controller compositional generalization experiment on the ant-pinpad environment.}}
    \label{tab:HparamsControllerCompAnt}
\end{table}
}

\newcommand{\HparamsControllerUnsupGrid}{
\begin{table}[htb]
    \centering
    \footnotesize
    \begin{tabularx}{\columnwidth}{X l}
        \toprule
        \textbf{Hyperparameter} & \textbf{Search} \\
        \midrule
        Observation coefficient $\lambda$ & 0 \\
        KL strength $\alpha$ & [0,0.05,0.1,0.17,0.3,0.5,1] \\
        Controller model & Low-rank ($16$) linear\\
        Controlled layer $l$ & $\frac{L}{2}$ \\
        Latent code dimension $n_z$  & 8 \\
        Controller Encoder hidden layer  & 64 \\
        Controller Decoder hidden layer  & 32 \\
        GRU dimension $n_h$ & 32 \\ 
        Sequence embedding dimension $n_s$ & 32 \\ 
        Train steps & 64000 \\
        Batch size & 512 \\
        Optimizer & AdamW \\
        \hspace{3mm} Learning rate & [1e-3] \\
        \hspace{3mm} Weight decay & [0.03] \\
        \hspace{3mm} $\beta$s & (0.9, 0.999) \\
        Scheduler \\
        \hspace{3mm} Constant learning rate \\
        \bottomrule
    \end{tabularx}
    \caption{\textbf{Hyperparameters for unsupervised abstract action discovery on the gridworld environment. }}
    \label{tab:HparamsControllerUnsupGrid}
\end{table}
}

\newcommand{\HparamsControllerUnsupAnt}{
\begin{table}[htb]
    \centering
    \footnotesize
    \begin{tabularx}{\columnwidth}{X l}
        \toprule
        \textbf{Hyperparameter} & \textbf{Search} \\
        \midrule
        Observation coefficient $\lambda$ & 0 \\
        KL strength $\alpha$ & [0,0.05,0.1,0.17,0.3,0.5,1] \\
        Controller model & Linear\\
        Controlled layer $l$ & $\frac{L}{2}$ \\
        Latent code dimension $n_z$  & 8 \\
        Controller Encoder hidden layer  & 64 \\
        Controller Decoder hidden layer  & 32 \\
        GRU dimension $n_h$ & 32 \\ 
        Sequence embedding dimension $n_s$ & 32 \\ 
        Train steps & 32000 \\
        Batch size & 128 \\
        Optimizer & AdamW \\
        \hspace{3mm} Learning rate & [3e-4] \\
        \hspace{3mm} Weight decay & [0.03] \\
        \hspace{3mm} $\beta$s & (0.9, 0.999) \\
        Scheduler \\
        \hspace{3mm} Constant learning rate \\
        \bottomrule
    \end{tabularx}
    \caption{\textbf{Hyperparameters for unsupervised abstract action discovery on the ant-pinpad environment.}}
    \label{tab:HparamsControllerUnsupAnt}
\end{table}
}

\newcommand{\HparamsControllerUnsupCompileGrid}{
\begin{table}[htb]
    \centering
    \footnotesize
    \begin{tabularx}{\columnwidth}{X l}
        \toprule
        \textbf{Hyperparameter} & \textbf{Search} \\
        \midrule
        MLP hidden dim $n_h$ & 32 \\
        Observation coefficient $\lambda$ & 0 \\
        KL strength for latent $z$ $\alpha_z$ & [0.003,0.01,0.03,0.1,0.3,1] \\
        KL strength for latent $\beta$ $\alpha_\beta$ & [0.003,0.01,0.03,0.1,0.3,1] \\
        Gumbel softmax temperature for $\beta$ & [0.5, 1] \\
        Maximum number of segments $M$ & 4 \\
        Prior switching rate & 10 \\
        Controller model & Low-rank (16) linear\\
        Controlled layer $l$ & $\frac{L}{2}$ \\
        Latent code dimension $n_z$  & 8 \\
        Train steps & 32000 \\
        Batch size & 512 \\
        Optimizer & AdamW \\
        \hspace{3mm} Learning rate & [1e-3] \\
        \hspace{3mm} Weight decay & [0.03] \\
        \hspace{3mm} $\beta$s & (0.9, 0.999) \\
        Scheduler \\
        \hspace{3mm} Constant learning rate \\
        \bottomrule
    \end{tabularx}
    \caption{\textbf{Hyperparameters for CompILE training on the gridworld environment.}}
    \label{tab:HparamsControllerUnsupCompileGrid}
\end{table}
}

\newcommand{\HparamsControllerUnsupCompileAnt}{
\begin{table}[htb]
    \centering
    \footnotesize
    \begin{tabularx}{\columnwidth}{X l}
        \toprule
        \textbf{Hyperparameter} & \textbf{Search} \\
        \midrule
        MLP hidden dim $n_h$ & 32 \\
        Observation coefficient $\lambda$ & 0 \\
        KL strength for latent $z$ $\alpha_z$ & [0.003,0.01,0.03,0.1,0.3,1] \\
        KL strength for latent $\beta$ $\alpha_\beta$ & [0.003,0.01,0.03,0.1,0.3,1] \\
        Gumbel softmax temperature for $\beta$ & [0.5, 1] \\
        Maximum number of segments $M$ & 4 \\
        Prior switching rate & 10 \\
        Controller model & Linear\\
        Controlled layer $l$ & $\frac{L}{2}$ \\
        Latent code dimension $n_z$  & 8 \\
        Train steps & 32000 \\
        Batch size & 128 \\
        Optimizer & AdamW \\
        \hspace{3mm} Learning rate & [3e-4] \\
        \hspace{3mm} Weight decay & [0.03] \\
        \hspace{3mm} $\beta$s & (0.9, 0.999) \\
        Scheduler \\
        \hspace{3mm} Constant learning rate \\
        \bottomrule
    \end{tabularx}
    \caption{\textbf{Hyperparameters for CompILE training on the ant-pinpad environment.}}
    \label{tab:HparamsControllerUnsupCompileAnt}
\end{table}
}

\newcommand{\HparamsInternalRLGrid}{
\begin{table}[htb]
    \centering
    \footnotesize
    \begin{tabularx}{\columnwidth}{X l}
        \toprule
        \textbf{Hyperparameter} & \textbf{Search} \\
        \midrule
        Policy model & SSM \\
        Policy depth & 1 \\
        Policy Embedding dimension & 256 \\
        Train steps & 100000 \\
        Batch size & 1024 \\
        Entropy regularizer & 0 \\
        Optimizer & AdamW \\
        \hspace{3mm} Learning rate & [3e-5] \\
        \hspace{3mm} Weight decay & [0.0] \\
        \hspace{3mm} $\beta$s & (0.9, 0.999) \\
        Scheduler \\
        \hspace{3mm} Constant learning rate \\
        \bottomrule
    \end{tabularx}
    \caption{\textbf{Hyperparameters for internal RL on the gridworld environment.}}
    \label{tab:HparamsInternalRLGrid}
\end{table}
}

\newcommand{\HparamsInternalRLAnt}{
\begin{table}[htb]
    \centering
    \footnotesize
    \begin{tabularx}{\columnwidth}{X l}
        \toprule
        \textbf{Hyper Parameter} & \textbf{Search} \\
        \midrule
        Policy model & SSM \\
        Policy depth & 1 \\
        Policy Embedding dimension & 256 \\
        Train steps & 51200 \\
        Batch size & 256 \\
        Entropy regularizer & 0 \\
        Optimizer & AdamW \\
        \hspace{3mm} Learning rate & [3e-5] \\
        \hspace{3mm} Weight decay & [0.0] \\
        \hspace{3mm} $\beta$s & (0.9, 0.999) \\
        Scheduler \\
        \hspace{3mm} Constant learning rate \\
        \bottomrule
    \end{tabularx}
    \caption{\textbf{Hyperparameters for internal RL on the ant-pinpad environment.}}
    \label{tab:HparamsInternalRLAnt}
\end{table}
}

\newcommand{\HparamsSSM}{
\begin{table}[htb]
    \centering
    \footnotesize
    \begin{tabularx}{\columnwidth}{X l}
        \toprule
        \textbf{Hyperparameter} & \textbf{Value} \\
        \midrule
        Embedding dimension $n_e$ & 256 \\
        Hawk LRU dimension & 256 \\
        Number of heads & 8 \\
        Variance scaling of all initializers & 0.1 \\
        MLP hidden layer dimension & 512 \\
        MLP nonlinearity & ReLU \\
        \bottomrule
    \end{tabularx}
    \caption{\textbf{Hyperparameters for Hawk state space model layers.}}
    \label{tab:HparamsSSM}
\end{table}
}

\newcommand{\HparamsTF}{
\begin{table}[htb]
    \centering
    \footnotesize
    \begin{tabularx}{\columnwidth}{X l}
        \toprule
        \textbf{Hyperparameter} & \textbf{Value} \\
        \midrule
        Embedding dimension $n_e$ & 256 \\
        Attention head dimension & 64 \\
        Number of heads & 4 \\
        Variance scaling of all initializers & 0.1 \\
        MLP hidden layer dimension & 512 \\
        MLP nonlinearity & ReLU \\
        Number of buckets for relative positional encodings & 32 \\
        \bottomrule
    \end{tabularx}
    \caption{\textbf{Hyperparameters for Transformer model layers.}}
    \label{tab:HparamsTF}
\end{table}
}

\usepackage[font=small,labelfont=bf]{caption}

\begin{abstract}
\small
Large-scale autoregressive models pretrained on next-token prediction and finetuned with reinforcement learning (RL) have achieved unprecedented success on many problem domains. During RL, these models explore by generating new outputs, one token at a time. However, sampling actions token-by-token can result in highly inefficient learning, particularly when rewards are sparse. Here, we show that it is possible to overcome this problem by acting and exploring within the internal representations of an autoregressive model. Specifically, to discover temporally-abstract actions, we introduce a higher-order, non-causal sequence model whose outputs control the residual stream activations of a base autoregressive model. On grid world and MuJoCo-based tasks with hierarchical structure, we find that the higher-order model learns to compress long activation sequence chunks onto internal controllers. Critically, each controller executes a sequence of behaviorally meaningful actions that unfold over long timescales and are accompanied with a learned termination condition, such that composing multiple controllers over time leads to efficient exploration on novel tasks. We show that direct internal controller reinforcement, a process we term ``internal RL'', enables learning from sparse rewards in cases where standard RL finetuning fails. Our results demonstrate the benefits of latent action generation and reinforcement in autoregressive models, suggesting internal RL as a promising avenue for realizing hierarchical RL within foundation models.

\end{abstract}

\title{Emergent temporal abstractions in autoregressive models enable hierarchical reinforcement learning}

\begin{document}
\maketitle

\noindent We are witnessing a revolution in artificial intelligence (AI), driven primarily by autoregressive sequence models. These models, most often built with transformers \citep{vaswani_attention_2017}, are trained using self-supervised next-token prediction on datasets of unprecedented scale \citep{kaplan_scaling_2020}. After pretraining, finetuning autoregressive models with reinforcement learning (RL) yields agents with competence in a wide range of domains and tasks, from mathematical problem solving, to being helpful assistants in scientific and creative human endeavors. Currently, there is great interest in leveraging RL as a means to discover new intelligent behaviors, beyond those present in the original training data \citep{guo_deepseek-r1_2025}.

RL efficiency can be greatly increased by starting from an autoregressive sequence model that has been pretrained on a wide range of behaviors, such as a large language model (LLM). From an RL standpoint, self-supervised pretraining can be seen as imitation learning under partial observability, where not only is noise introduced and intermediate steps occluded, but also latent variables, such as task descriptors, agent rewards and goals, and their mental states, are unknown. This setup imbues the resulting models with latent variable inference capabilities \citep{xie_explanation_2022,oswald_uncovering_2023} (commonly referred to as in-context learning \citep{brown_language_2020}) that allow adapting to new tasks and environments quickly. Moreover, pretrained autoregressive models serve as rich action priors from which diverse, meaningful sequences can be sampled, enabling efficient exploration from the start.

\begin{figure*}[t!]
        \centering
    
        \includegraphics[width=.97\textwidth]{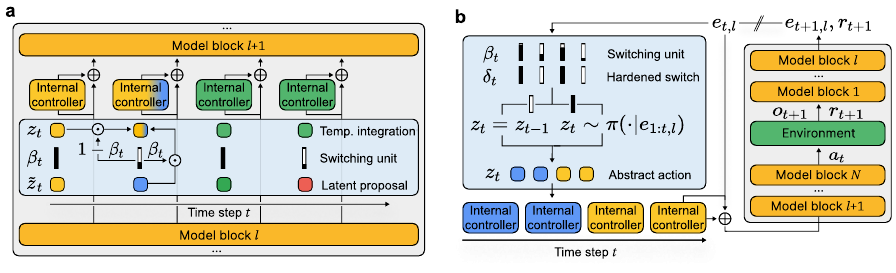}\\[-\dp\strutbox]

        \caption{\textbf{Research overview.} (\textbf{a}) We let a metacontroller steer the residual stream activations of a pretrained autoregressive model. Through self-supervised next-action prediction, the metacontroller discovers how to generate sequences of simple (linear) internal controllers that change sparsely in time, following a dynamic switching unit $\beta_t \in [0,1]$. In hierarchically-structured tasks, each internal controller corresponds to a temporally-abstract action that leads the base autoregressive model to achieve a meaningful elementary goal. (\textbf{b}) We perform RL internally -- in the abstract space discovered by the metacontroller -- by subsuming the autoregressive model into the environment and acting in the residual stream on a contracted timescale.\label{fig:overview}}
\end{figure*}
Efficient, long-horizon exploration is key for RL to succeed, in particular when rewards are sparse. This leads us to an important problem that autoregressive models face: because these models produce sequences one token at a time, RL exploration is driven entirely by token-level variations. However, solely relying on token-by-token variability to explore can be insufficient to make progress on hard, sparse-reward problems which require generating multiple tokens correctly before obtaining a reward. This observation, which is at the center of the present study, has motivated a long line of research on hierarchical RL. Hierarchical RL attempts to exploit the fact that real-world problems are typically amenable to a hierarchical approach, wherein a final solution is expressed in terms of temporally-abstract actions --- i.e., reusable subroutines that run for extended time periods (sometimes called ``options'') \citep{sutton_between_1999}. Evidence suggests that humans approach problem solving using such temporal abstractions \citep{botvinick_hierarchically_2009}, which implies that this may be a very efficient way to learn. Importantly, if temporally-abstract subroutines exist, exploration can occur at higher levels of temporal abstraction, drastically reducing the search space relative to token-by-token exploration. However, discovering appropriate subroutines via deep RL remains a longstanding challenge. While policy gradient methods have been derived (e.g., the option-critic \citep{bacon_option-critic_2017}), these approaches have theoretical issues and tend to fail in practice, often converging to degenerate options \citep{pateria_hierarchical_2021}.

In this paper, we pursue an alternative approach for temporally-abstract action discovery that builds directly upon autoregressive modeling. Based on their in-context latent variable inference capabilities, we hypothesize that autoregressive action models implicitly learn temporally-abstract actions represented in their internal activations, despite being trained to predict only one token at a time. This hypothesis leads us to introduce an internal neural network controller in charge of steering the internal activations of a base model. Critically, the controller learns through an unsupervised variational inference algorithm \citep{peterson_mean_1987,hinton_keeping_1993,kingma_auto-encoding_2014,rezende_stochastic_2014}, which does not require per-time-step abstract action labels, in contrast to standard model steering techniques \citep{zou_representation_2023,turner_steering_2023}.

We evaluate our approach on a family of RL tasks that are constructed in a hierarchical, compositional manner. We consider both a classic discrete grid world environment \citep{kipf_compile_2019,jiang_learning_2022}, and a more challenging hierarchical continuous control environment implemented on the MuJoCo physics simulator \citep{todorov_mujoco_2012}. The latter requires an agent to master both low-level continuous motor control as well as planning at a higher level of temporal abstraction to exploit the underlying discrete, compositional task structure.
We find that the internal controller discovers how to generate higher-order sequences of temporally-abstract actions that switch sparsely in time. These abstract actions enable efficient exploration by drastically reducing the search space size in novel tasks and simplify credit assignment by reducing the effective time horizon of the policy. The final product is a novel hierarchical RL method that directly reinforces internal activations to solve sparse reward tasks that token-level approaches cannot solve. Our results demonstrate the benefits of latent action generation for RL applied to pretrained autoregressive models.

\subsection*{Key results}
We illustrate our approach in Fig.~\ref{fig:overview}, and preview our main contributions below:

\begin{itemize}
\item \textbf{Next-action predictors inherently develop temporally-abstract action representations.} We analyze transformers and state-space models (SSMs) trained to autoregressively predict the actions of goal-directed agents, whose goals are unknown. We find that the networks learn to represent (and infer in-context) a belief about an agent's goals in their residual stream activations.

\item \textbf{Linearly controllable temporally-abstract actions.} These temporally-abstract representations are also easily controllable: a linear residual stream controller near mid-depth suffices to turn the sequence model into a closed-loop goal-optimizing policy, capable of executing a long-horizon plan.

\item \textbf{Compositional generalization in the residual stream.} We show that such controllers can be sequenced in time. Residual stream controller sequencing enables compositional generalization, yielding agents that combine multiple goals in ways not seen during training.

\item \textbf{A new neural architecture for autoregressive model control, which discovers temporally-abstract actions without supervision.} We develop a metacontroller neural network that reads from the sequence model residual stream, and in return applies a linear controller to it. The metacontroller learns to generate goal-optimizing controllers that exhibit temporal abstraction: it keeps applying the same controller for a variable number of time steps before switching to a new one. To discover appropriate temporally-abstract actions without any supervision signals, our method relies on two key properties: (i) reading from and writing back to the residual stream of a pretrained autoregressive model, and (ii) future-conditioning: during training, the metacontroller is non-causal, and is conditioned on a sequence embedding obtained by performing a first pass through the entire sequence.

\item \textbf{A new ``internal RL'' paradigm, many orders of magnitude faster than standard RL finetuning in hierarchically-structured tasks.} We introduce internal RL: performing RL directly within the residual stream of the base model, taking internal activations as observations and metacontroller outputs as actions. We show that internal RL significantly outperforms both standard RL finetuning as well as a strong prior hierarchical RL method \citep[CompILE;][]{kipf_compile_2019}, achieving both higher initial success rates and more efficient credit assignment than the baseline methods in hierarchically-structured tasks.

\end{itemize}

\section*{Results}

\subsection*{Linearly controllable abstract action representations emerge in autoregressive models}
Before diving into the description of our internal RL model, we first analyze the internal activations of autoregressive models pretrained to predict the behavior of goal-directed agents. Our goal here is to verify that a model trained on next-token prediction can learn temporally-abstract actions in its internal activations that we can leverage for internal RL. To do this, we pretrain our models from scratch on a behavioral dataset $D$ comprising observation-action sequences produced by different expert agents that solve tasks via stochastic policies of varying degrees of optimality. The autoregressive model can thus be thought of as a sequence model of likely observation-action trajectories. Each element of $D$ is a sequence $(o_1, a_1, \ldots, a_T, o_{T+1})$ comprised of the initial sensory observations $o_1$, actions $a_t$ taken by an agent and resulting sensory observation $o_{t+1}$ at time steps $t \in \{1,\ldots,T\}$. Like behavioral datasets collected at scale (e.g., those used to train LLMs), $D$ does not contain rewards, nor any explicit agent goal and task descriptors. The analyses presented in this section seek to determine if, and how, autoregressive models infer abstract patterns in long-horizon, goal-directed action sequences.

\begin{figure}[h!]
    \centering
    \begin{minipage}{.5\textwidth}
    \includegraphics[width=\textwidth]{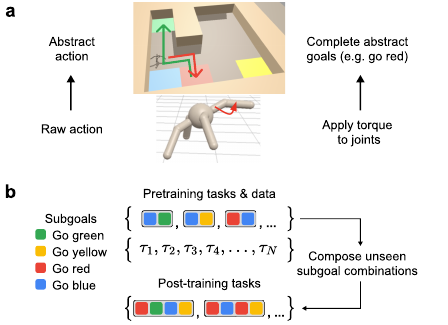}
    \end{minipage}
    \caption{\textbf{Environment and task design.} (\textbf{a}) To complete a task, an agent must visit in sequence a number of subgoal locations, each marked with a specific color. The tasks are performed either in a discrete grid world or in a continuous motor control environment, illustrated above, where a quadrupedal robot (the `ant') must be actuated at joint level. A task can be described as an abstract action sequence (the subgoal locations that must be visited), or as a sequence of low-level motor commands. (\textbf{b}) We pretrain autoregressive action models and metacontrollers on unlabeled behavioral datasets containing observation-action sequences of expert agents performing different tasks. These sequences do not contain rewards or subgoal labels. We then test the ability of the models to learn with RL tasks that comprise longer subgoal sequences, combined in new orders not seen during pretraining and metacontroller training.}
    \label{fig:task-hierarchy}
\end{figure}

We collect behavior from two classes of environments where agents perform navigation tasks. Importantly, the tasks are hierarchically-structured (cf.~Fig.~\ref{fig:task-hierarchy}): though basic movement skills are a prerequisite, any given task can be solved with a combination of sub-routines composed of common sequences of basic movements. More concretely, we study both a discrete grid world environment that was previously introduced as a testbed for hierarchical RL \citep{kipf_compile_2019,jiang_learning_2022}, as well as a continuous-observation, continuous-action adaptation implemented by us in the MuJoCo physics simulator \citep{todorov_mujoco_2012}, where a quadrupedal robot (the `ant' \citep{schulman_high-dimensional_2015,fu_d4rl_2020}) must be controlled at joint-level. In both environments, an agent needs to follow a course that arrives at certain colored locations in a specific order. In other words, the agents need to navigate between subgoals while also ignoring distractors (non-goal colored locations), all while avoiding collisions with randomly placed walls. Any task is described by a sequence of subgoals, which are either a single colored location for the ant, or two consecutive colored locations for the grid world. A given task can be mapped to different spatial configuration of the subgoals, the distractors, and the walls, see Appendix~\ref{app:environmental_details} for more details on the environments. In these environments, abstract actions are equivalent to moving towards a specific subgoal, hence we use the terms ``abstract action'' and ``subgoal'' interchangeably in this paper.

Given behavioral data collected for a set of easy tasks, referred to as pretraining tasks set (see Appendix ~\ref{app:environmental_details} and~\ref{app:pretraining_details} for more details on the tasks and how the behavioral data are collected), we proceed with autoregressive sequence model pretraining, here a standard causal transformer \citep{vaswani_attention_2017} for discrete grid world data, and an efficient SSM (Hawk \citep{de_griffin_2024}) for ant control data. The models are pretrained from scratch by minimizing the cross-entropy
\begin{equation}
    L(\theta) = \sum_{(o_{1:T+1}, a_{1:T}) \sim D} \sum_{t=1}^{T} -\ln p_{\theta}(a_{t}|o_{1:t}) - \lambda \ln p_{\theta}(o_{t+1}|o_{1:t}),\nonumber
\end{equation}
with $p_\theta$ the sequence model, and $\theta$ its parameters. For the case of continuous actions, the likelihood $p_{\theta}(a_{t}|o_{1:t})$ is modeled as a Gaussian with learned diagonal covariance matrix. For discrete actions, the likelihood is parameterized as a categorical distribution with probabilities provided by the softmax over the output logits. Note that while the main objective here is behavioral (next-action) prediction, the models are also trained on next-observation prediction, the objective of world (dynamics) modeling \citep{schmidhuber_making_1990,sutton_dyna_1991,hafner_mastering_2025}. The weight of this auxiliary loss is determined by a scalar hyperparameter $\lambda \ge 0$; we analyze its role in the Appendix Fig.~\ref{appfig:wd_obs_scaling}. Additional optimization and architectural details may be found in Appendix~\ref{app:experimental_details} and~\ref{app:architecture}.
\begin{figure}[ht]
        \centering
        \setlength{\lineskip}{0pt}
        \setlength{\parskip}{0pt}
    
        \includegraphics[width=0.485\textwidth]{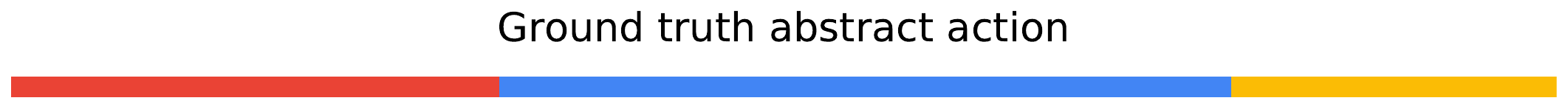}\\[-\dp\strutbox]
        \vspace{0.4cm}
        \includegraphics[width=0.485\textwidth]{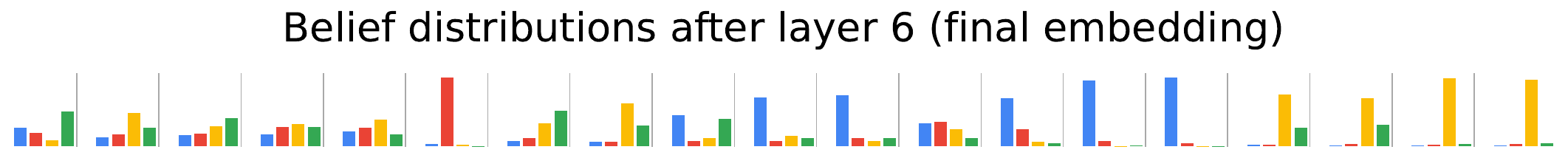}\\[-\dp\strutbox]
        \vspace{0.4cm}
        \includegraphics[width=0.485\textwidth]{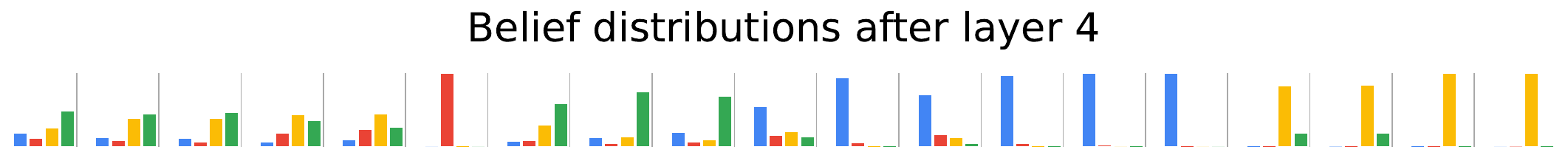}\\[-\dp\strutbox]
        \vspace{0.4cm}
        \includegraphics[width=0.485\textwidth]{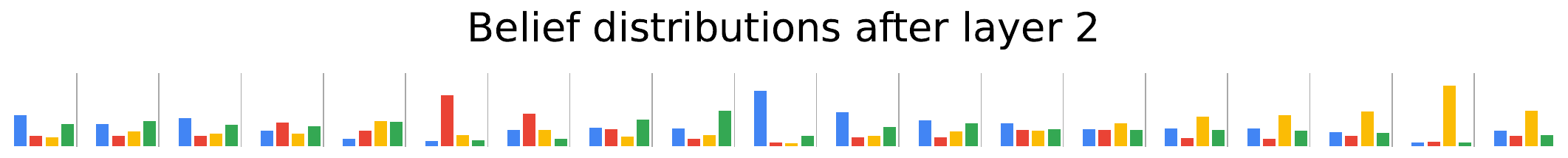}\\[-\dp\strutbox]
        \vspace{0.4cm}
        \includegraphics[width=0.485\textwidth]{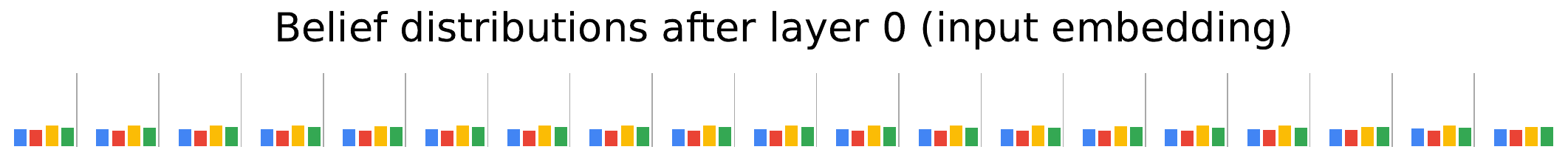}\\[-\dp\strutbox]
        \vspace{0.15cm}
        \includegraphics[width=0.485\textwidth]{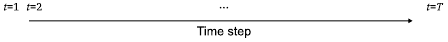}\\[-\dp\strutbox]

        \caption{\textbf{Internal belief distributions over abstract actions, according to a linear probe.} Decoding performance of linear classifiers trained to predict groundtruth abstract actions from instantaneous residual stream activation vectors increases until mid-depth (layer 4) and remains strong up until the final embedding vector $e_{t,L}$ (here $L=6)$. \label{fig:belief-state-probing}}
\end{figure}

To determine whether the internal activations of the pretrained autoregressive models learn to identify temporal abstractions related to the subgoals, we analyze the internal activations of the models using two common mechanistic interpretability techniques \citep{sharkey_open_2025}, linear probing \citep{alain_understanding_2017} and causal model intervention \citep{geiger_causal_2021,meng_locating_2022}. For the former (linear probing), we train linear classifiers to decode the agent subgoals $g_t\in\{1,\ldots,G\}$ on the grid world environment from the instantaneous (time step $t$) residual stream activation vector $e_{l,t} \in \mathbb{R}^{n_e}$ after the $l$-th model block. Fig.~\ref{fig:belief-state-probing} shows that linear decoder probability mass concentrates on the correct latent subgoal as time $t$ increases, i.e. as more evidence about the current agent subgoal is gathered. Moreover, linear decoding likelihood increases with layer depth $l$, peaking close to the final embedding used by the transformer decoder. Thus, despite being trained only on one-step action prediction, the autoregressive models learn to represent temporally-abstract subgoals. This result is in line with the infinite-data theory of in-context Bayesian inference in sequence predictors \citep{ortega_meta-learning_2019}, and adds more evidence to the linear representation hypothesis in neural sequence models \citep{nanda_emergent_2023,park_linear_2024,gurnee_language_2023}.

For causal model intervention, we ask whether the internal representations of the autoregressive model can be leveraged to create a subgoal-optimizing policy. Inspired by the effectiveness of LoRA finetuning \citep{hu_lora_2022}, we introduce a low-rank linear residual stream controller with parameters $U \in \mathbb{R}^{n_e \times n_e}$, which modifies the instantaneous residual stream activations in between model blocks at a given depth $l$ following the update
\begin{equation}
\label{eq:additive-controller}
e_{t,l} \gets e_{t,l} + U_t e_{t,l}.
\end{equation}
Note that we allow the controller parameters $U_t$ to vary in time. In this section, we maintain a set of $G$ separate controllers $\{U^{(g)}\}_{g=1}^{G}$, one per subgoal, and manually select which controller $U_t$ to apply at every time step $t$ using the groundtruth subgoal label $g_t$. (We will eliminate the use of ground-truth subgoal labels later on.) To train the controllers, we condition generation upon the correct subgoal-specific controller $U^{(g)}$, and minimize the cross-entropy $\sum_{(o_{1:T+1}, a_{1:T}) \sim D_*} \sum_t -\ln p_{\theta,\phi}(a_{t}|o_{1:t}, g_t)$ w.r.t.~controller parameters $\phi$ (while holding $\theta$ fixed) on a behavioral dataset $D_*$. This dataset contains behavioral sequences that are generated in the same way as those in the pretraining dataset $D$, but with increased optimality, see Appendix~\ref{app:metacontroller_demo}. Here and throughout, $\phi$ refers to controller parameters that were not part of the pretrained model $p_\theta$, and $p_{\theta,\phi}$ denotes a controlled model.

\begin{figure}[b!]
    \centering
    \begin{minipage}{.48\textwidth}
    \includegraphics[width=\textwidth]{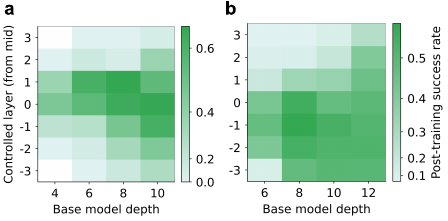}
    \end{minipage}
    \caption{\textbf{Mid-depth linear internal controllers achieve length and compositional generalization.} Both panels analyze success rate (the fraction of rewarded trials in which the full sequence of elementary goals defining a given task is completed) as a function of base model depth (the number of autoregressive model layers) and controlled layer (the layer at which the internal controller is inserted, with 0 corresponding to the middle of the base model). In both grid world (\textbf{a}) and ant (\textbf{b}) environments, inserting the controller near the middle layer results in better controllability, as measured by the success rate on the post-training tasks, which require both length and compositional generalization. To produce this analysis, we trained one controller per subgoal using groundtruth labels; to evaluate success rates we activated the controllers in correct order, again using groundtruth subgoal labels. Results averaged over 5 seeds.
    \label{fig:control-depth-analysis}}
\end{figure}

We evaluate the subgoal-optimizing controllers on a post-training OOD task set that requires both length and compositional generalization: as shown in Fig.~\ref{fig:task-hierarchy} and detailed in Appendix~\ref{app:environmental_details}, the post-training tasks recombine subgoals in orders not seen neither during pretraining nor controller training. As well, they comprise longer subgoal trajectories. Fig.~\ref{fig:control-depth-analysis} shows that these novel tasks can be solved with a high success rate by simply activating the corresponding subgoal controllers in the correct order, without any autoregressive sequence model retraining. More detailed descriptions of these mechanistic interpretability experiments and some additional experimental results are presented in Appendix~\ref{app:additional_exp} and~\ref{app:experimental_details}.

Our analysis further reveals a distinction between latent variable belief state representation (at least w.r.t.~a linear decoder) and internal representation control. Whereas linear subgoal decoding is possible from mid-depth up until the final layer, subgoal-conditioning is best achieved by inserting a linear controller in the middle of the pretrained sequence model, see Fig.~\ref{fig:control-depth-analysis}. There is an intuitive appeal to this result: the mapping from abstract subgoals spanning many time steps to actual per-time-step low-level actions is implemented over multiple model layers.
Our findings join two recent studies \citep{skean_layer_2025,csordas_language_2025} that identify the first half of language models as the strongest for transfer learning, and as exerting the strongest influence on predicting future tokens. Given these results, in what follows, and unless noted otherwise, controllers always read from and write back to the residual stream at mid-depth of the autoregressive sequence model.

\subsection*{Unsupervised metacontroller discovers temporally-abstract actions within autoregressive models}

The analyses above show that simple internal activation controllers can steer a pretrained next-action sequence model to execute temporally-abstract actions, here navigation to a sequence of subgoals. We have so far assumed access to subgoal labels, similarly to how current model steering methods \citep{wu_axbench_2025} are trained using detailed supervision information (e.g., on the truthfulness of an answer \citep{li_inference-time_2022} or on personality traits \citep{chen_persona_2025}). We now turn to the challenging unsupervised setting with no groundtruth labels, where the model must both discover temporally-abstract actions from an unlabeled behavioral dataset $D$, and learn a selection mechanism that generates appropriate sequences of subgoals, and related abstract actions, in order to achieve a larger goal.

\begin{figure}[htbp!]
    \centering
    \includegraphics[width=0.95\linewidth]{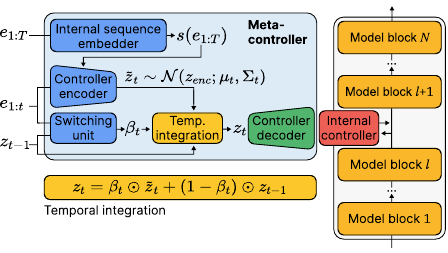}
    \caption{\textbf{Details of the metacontroller architecture and the different modules at play.} The metacontroller learns in a self-supervised way to generate sequences of internal controllers. Candidate controller codes $\tilde{z}_t$ are sampled from a Gaussian with context-dependent mean and covariance, and are integrated at a continuous, time-varying rate $\beta_t$, dynamically determined by a switching unit. Values of $\beta_t$ close to zero ignore new controller candidates; conversely, values close to unity lead to switching to a new controller. This mechanism is key for achieving temporal abstraction. The metacontroller features another key design element, a future-conditioned encoder: during self-supervised learning, the metacontroller is non-causal, and has access to the entire sequence of residual stream activations through a sequence embedding $s(e_{1:T})$.}
    \label{fig:metacontroller-arch}
\end{figure}

To simultaneously learn abstract actions and orchestrate their execution, we freeze the autoregressive model after training on $D$, then we augment it with a metacontroller that can generate the controllers, $U_t$, for the residual stream activations in the sequence model. As before, we continue training on $D_*$ with $\theta$ fixed. But, now, we do not condition the controller on the groundtruth subgoal --- instead the metacontroller learns how to generate the appropriate controllers at the appropriate times. 
We describe the model in full in Appendix~\ref{app:meta_controller}, and illustrate it in Fig.~\ref{fig:metacontroller-arch}. Briefly, the metacontroller is a generative stochastic recurrent neural network with an encoder-decoder architecture that enables sampling controllers sequentially. Because it outputs the parameters $U_t$ of a controller and not directly a control vector, the metacontroller can be qualified as a recurrent hypernetwork \citep{ha_hypernetworks_2017}. The decoder is a feedforward network that produces a controller, $U_t$, from a controller code, $z_t$. The encoder is a recurrent network based on the gated recurrent unit \citep[][]{cho_properties_2014} that specifies the mean $\mu_t$ and variance $\Sigma_t$ of a Gaussian distribution over a random controller code $\tilde{z}_t \sim \mathcal{N}(z_\text{enc}; \mu_t, \Sigma_t)$. Importantly, the encoder is non-causal, because it receives an embedding, $s(e_{1:T})$, of the whole sequence of latent activities. We justify such future-conditioning using a formal latent variable modeling argument in Appendix~\ref{app:elbo}.

Additionally, the metacontroller includes a recurrent switching unit, that operates between the encoder and decoder. This unit determines a time-varying continuous switching gate $\beta_t \in [0, 1]$, which controls the interpolation between previous controller code $z_{t-1}$ and a new sampled code $\tilde{z}_t$:
\begin{equation}
\label{eq:switching-unit}
    z_t = \beta_t \odot \tilde{z}_t + (1-\beta_t)\odot z_{t-1},
\end{equation}
where $\odot$ denotes elementwise multiplication. Despite its simplicity, this temporal integrator is critical for the metacontroller to learn to generate the appropriate temporally-abstract actions, as we will confirm through ablation experiments at the end of this section.

The metacontroller parameters $\phi$ are trained through the minimization of a self-supervised learning objective, comprising (low-level) next-action prediction and an additional prior-matching regularizer,
\begin{equation}
\begin{aligned}
\label{eq:metacontroller-loss}
    L(\phi) = & \sum_{(o_{1:T+1}, a_{1:T}) \sim D_*} \sum_{t=1}^{T} \Big[ -\ln p_{\theta,\phi}(a_{t}|o_{1:t}, z_{1:t}) \\
    &+ \alpha D_{\mathrm{KL}}\big(\mathcal{N}(\mu_t, \Sigma_t) \,\|\, \mathcal{N}(0, I)\big)\Big],
\end{aligned}
\end{equation}
where $D_\text{KL}(\cdot \,\|\, \cdot)$ denotes the Kullback-Leibler divergence \citep{cover_elements_2006}. The inclusion of this regularizer (with weight determined by the hyperparameter $\alpha \ge 0$) promotes the generation of meaningful sequences when sampling controller codes $z_t$ from a standard normal distribution, a property that we exploit in the next section to develop a novel hierarchical RL algorithm. From an information-theoretic perspective, $\alpha$ also controls the variational bottleneck by regulating the information flow from the acausal encoder to the controller. As shown in our later analysis, this bottleneck is instrumental in driving the model toward sparse, subgoal-aligned switching patterns that mirror the underlying task structure. Moreover, the choice of an unconditional prior (i.e., where next abstract action proposals are independent of past ones) promotes the development of compositional representations, which match well our hierarchical tasks. In Appendix~\ref{app:elbo}, we derive \cref{eq:metacontroller-loss} formally using a variational information-theoretic approach \citep{alemi_fixing_2018}. The derivation is standard, and follows closely previous calculations for stochastic recurrent models \citep[e.g.,][]{linderman_bayesian_2017,kim_variational_2019}.

\begin{figure}[b!]
    \centering
    \hspace*{.15cm}\includegraphics[width=0.48\textwidth]
    {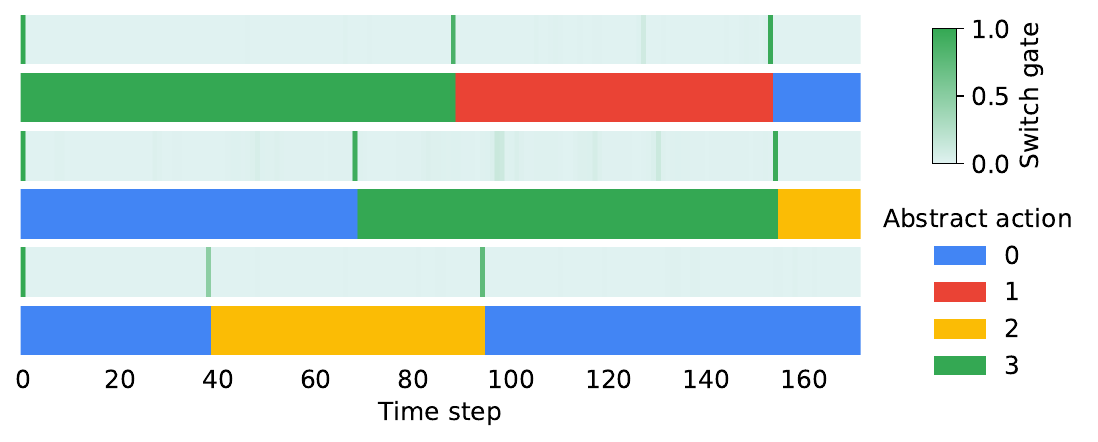}
    \caption{\textbf{Self-supervised metacontroller discovers temporally-abstract actions within pretrained autoregressive model.} Three example trajectories from the ant control environment showing the switch $\beta_t$ used for temporal integration at each timestep, and the groundtruth abstract action being performed (color-coded). Switching ($\beta_t \approx 1$) coincides with a change in the abstract action being performed. \label{fig:switch-gate-analysis}}
\end{figure}

Ultimately, the metacontroller both discovers the temporally-abstract actions that underlie the observed agents' behavior, and learns to sequence them appropriately in time by implementing respective termination conditions via the switching gate. In Fig.~\ref{fig:switch-gate-analysis} and~\ref{appfig:cossim_grid}, we analyze the residual stream controllers discovered by the metacontroller by plotting the switching gate values $\beta_t$ against groundtruth abstract actions $g_t$.
We find that the metacontroller recovers the groundtruth abstract action switching times. After training, the switch gate learns to behave in a quasi-binary, sparsely-switching fashion, despite not being explicitly regularized to do so. This is a notable finding in light of the critical role that switching regularization methods play in hierarchical RL \citep{harb_when_2018}, and given the simplicity of the temporal integrator (\cref{eq:switching-unit}). The resulting temporal segmentation is essentially perfect, despite the fact that both observations and actions are continuous for the ant environment. Moreover, the metacontroller learns to generate latent controller codes which correspond to meaningful temporally-abstract actions (e.g., ``go to color blue''), that generalize to new task configurations and switching times (see Appendix~\ref{app:abstract_action} for an analysis).

\begin{figure}[]
    \centering
    \hspace{-0.3cm}
    \hspace{0.1cm}\includegraphics[width=0.49\textwidth]{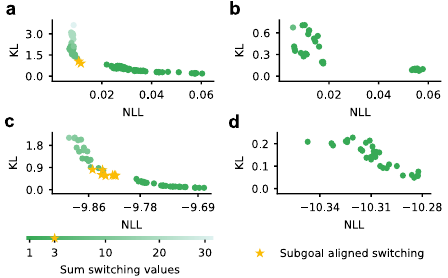}
    \caption{\textbf{A rate-distortion analysis reveals the importance of the controlled, pretrained autoregressive model being frozen for the discovery of temporally-abstract actions.} We compare our standard metacontroller, which steers a frozen base model (left column; \textbf{a}, \textbf{c}), with a metacontroller that is co-trained with the base model it is steering (right column; \textbf{b}, \textbf{d}). The x-axis represents action prediction loss (the distortion, or negative log-likelihood; NLL) and the y-axis represents the KL divergence to the prior (the rate). As the trade-off hyperparameter $\alpha$ in \cref{eq:metacontroller-loss} is swept over to trace the rate-distortion curve, it reveals a range of values for which correct subgoal switching representations develop (marked with a $\star$) when the base model is frozen, but not for the co-training regime.
    This holds similarly for grid world (top row; \textbf{a}, \textbf{b}) and the ant environment (bottom row; \textbf{c}, \textbf{d}).}\label{fig:rate_distortion_plot} 
\end{figure}

We next study what happens when the autoregressive base model parameters $\theta$ are not kept frozen, and instead co-trained with metacontroller parameters $\phi$ through variational inference (the minimization of \cref{eq:metacontroller-loss}, now w.r.t.~both $\theta$ and $\phi$). This baseline is conceptually close to previous hierarchical RL methods that use variational inference to learn abstractions from unlabeled demonstrations (e.g., \citep{kipf_compile_2019,jiang_learning_2022}), while using our particular neural network architecture. To compare the abstract action representations developed when the base model is frozen vs.~when it is not,  we resort to a rate-distortion analysis \citep{alemi_fixing_2018}, obtained by varying the value of the hyperparameter $\alpha$ (which controls the rate-distortion trade-off in \cref{eq:metacontroller-loss}) over a wide interval, see \ref{app:rate_distortion} for additional details. We trace rate-distortion curves for both our standard metacontroller (which steers a pretrained, frozen autoregressive model) and for the co-trained metacontroller, see Fig.~\ref{fig:rate_distortion_plot}.

Intriguingly, we find that a horizontal gap appears on the rate-distortion curve between metacontrollers with subgoal-aligned switching (with rate-distortion points marked by a $\star$ symbol in Fig.~\ref{fig:rate_distortion_plot}), and those with slightly less rate. This indicates that at that rate level, a small increase in rate dramatically improves the distortion. In contrast, for the co-trained metacontroller, although the variational objective is minimized, this structure is lost. For most values of $\alpha$, the model converges to a degenerate solution characterized by a single switch at the very beginning of the sequence. The fact that subgoal-aligned switching corresponds to this improved distortion with frozen autoregressive models, but not with co-trained models, shows that pretraining builds an internal representation that aligns well with abstract actions. Furthermore, this also has optimization implications: for a given value of $\alpha$, the variational objective (\cref{eq:metacontroller-loss}) is minimized on the point of the rate distortion curve which has a tangent of slope $-1/\alpha$. A gap like the above, with a slope discontinuity, indicates that for a large range of values of $\alpha$, the variational objective is minimized precisely at the region with subgoal-aligned switching. This analysis therefore confirms that controlling a frozen autoregressive action predictor is essential for the discovery of temporally-abstract actions.

Taken together, the results presented in this section provide strong evidence that our model can both learn temporally-abstract actions and how to sequence them appropriately, all in a self-supervised manner. We will see next how this model can be leveraged to speed up exploration in new, harder tasks by many orders of magnitude, enabling sparse-reward RL to succeed.

\subsection*{Internal reinforcement learning}
Finally, we consider the question of how to leverage our model to learn harder tasks through hierarchical RL. We study only the challenging sparse-reward setting, where a single positive success reward is provided per trajectory, and only when an entire sequence of subgoals is correctly completed.

\begin{figure}[htb]
    \centering
    \hspace{-0.3cm}
    \includegraphics[width=.49
    \textwidth]{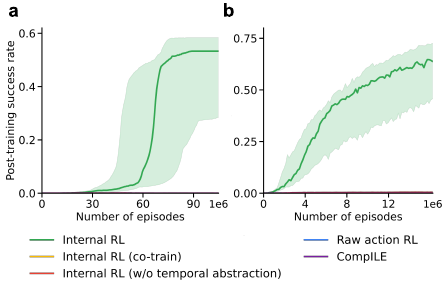}
    \caption{ \textbf{Internal reinforcement learning solves sparse-reward compositional tasks where standard methods fail.} RL curves for various methods that leverage a pretrained autoregressive sequence model for the (\textbf{a}) discrete grid world environment, and (\textbf{b}) the ant continuous control environment. We compare our full-blown internal RL algorithm to a number of baselines: standard (raw action) RL finetuning; CompILE \citep{kipf_compile_2019}, a hierarchical RL method that also learns from unlabeled demonstrations, like ours; internal RL applied to a metacontroller that has been trained without a temporal integration unit (forced switching at every timestep, $\forall_t \; \beta_t=1$); and internal RL applied to a metacontroller that has been co-trained from scratch with an autoregressive action model, sidestepping the pretraining phase (see main text for more details). All baselines fail to learn within a million episodes. Lines and shaded area resp.~report median and the spread between the $25^\text{th}$ and $75^\text{th}$ quantiles computed over 30 runs (3 metacontrollers trained for each of 10 pretrained models). We provide this figure in log-scale in Appendix Fig.~\ref{fig:internal-RL-logscale} for a more detailed analysis of the failure modes of the baselines.
    \label{fig:internal-RL}}
\end{figure}

We begin this section by establishing that our tasks (described in Fig.~\ref{fig:task-hierarchy}) are difficult for standard RL approaches to post-training. We first study an adapted version of the GRPO algorithm \citep{guo_deepseek-r1_2025}, which is a strong baseline in the sparse-reward setting. The details of our GRPO implementation can be found in Appendix~\ref{app:grpo}. For the tasks considered here, training an agent from scratch directly with RL has, for all practical purposes, no chance of succeeding.
Thus, to make the comparison fair, we instead apply GRPO to the pretrained autoregressive sequence model, as is now routinely done with LLMs. However, even with a pretrained sequence model that has been trained on action sequences related to the subgoals, there is only a minuscule chance (on the order of one in a million) of producing successful trajectories by random sampling at the output token-level. This causes GRPO training to fail, as the model does not receive enough signal to learn, see Fig.~\ref{fig:internal-RL}. An inspection of the action sequences generated by the autoregressive sequence model reveals that while the model reproduces action sequences seen in the training data, it fails to explore at a higher level of temporal abstraction, which would be required to solve these sparse reward RL tasks. In other words, simply training the sequence model with policy gradients does not lead the system to explore novel combinations of subgoals.

Having shown that standard post-training RL fails, we now introduce internal RL. The key step in internal RL is to treat the autoregressive sequence model as part of the environment; actions then correspond to residual stream interventions, $u_t$, and observations correspond to residual stream activations, $e_{t,l}$. We note that performing RL at the residual stream level is \emph{a priori} challenging. Consider the problem of learning from scratch a policy $\pi(u_t \mid e_{1:t})$ whose outputs $u_t \in \mathbb{R}^{n_e}$ additively control the residual stream, $e_{t,l} \gets e_{t,l} + u_t$, without relying on error backpropagation to differentiate through the base model that is being controlled. This is a high-dimensional continuous control problem, an exceedingly difficult setting for RL \citep{lillicrap_continuous_2016}.

Instead of directly attempting to learn a residual stream control policy, internal RL consists of doing RL in the controller code space of $z$, after the metacontroller is trained in a self-supervised manner, as described in the previous section. This approach assumes that the metacontroller has learned a meaningful switching unit $f_\text{switch}$, and a controller code space such that $z_t \sim \mathcal{N}(0, I)$ is a meaningful prior for sampling abstract actions. Intuitively, the metacontroller does not suffer from the drawbacks of directly doing RL in the residual stream for two reasons: (i) the action space dimension is reduced ($n_z < n_e$), (ii) the metacontroller operates on an abstract timescale, dramatically reducing the time horizon for difficult environments. The latter is the key property that can enable internal RL to be more efficient and succeed on hierarchical, sparse reward tasks where standard RL methods fail.

In more detail, internal RL consists in replacing an unsupervised controller encoder which uses privileged future information $s(e_{1:T})$ by a causal abstract action policy $\pi(z_t \mid e_{1:t})$, and then training it through RL, while keeping all other modules and their parameters fixed. Conceptually, this amounts to subsuming the autoregressive model, as well as part of the metacontroller, into the environment (cf. Fig.~\ref{fig:overview}). To generate discrete switching events, we further apply a threshold to binarize the switching rate, i.e., we replace $\beta_t$ in \cref{eq:switching-unit} by $H(\beta_t - \beta_\text{threshold})$ with $H$ the Heaviside step function and $\beta_\text{threshold} \in \mathbb{R}$ a hyperparameter.
This way, until a switch signal ($\beta_t=1$) is emitted by the metacontroller, the same abstract action is applied, thus allowing $\pi$ to operate on a temporally-abstract timescale. Pseudocode for the internal RL environment and algorithm is provided in Appendix~\ref{app:internal_rl}.

Fig.~\ref{fig:internal-RL} shows that internal RL achieves a high success rate on the post-training task set. Leveraging the temporal abstractions discovered through self-supervised metacontroller learning is crucial for this success, as shown by the failure of a metacontroller for which the temporal integration unit is disabled ($\forall_t \; \beta_t = 1$). To give this baseline a fair chance, this ablation is introduced during self-supervised metacontroller learning, not just when performing post-training RL. We note that the $\beta_t=1$ ablation also achieves a high initial success rate; this can be seen when plotting success rates in log-scale (cf.~Appendix~Fig.~\ref{fig:internal-RL-logscale}). However, only our full-blown (temporally-abstract) internal RL both achieves high initial success rates and performs efficient credit assignment, such that RL succeeds. In Appendix~\ref{app:internal_rl_vs_reparam} we present a mathematical argument for the efficiency of credit assignment in internal RL, comparing the variance of the resulting policy gradients of internal against RL in raw action space.

Moreover, to evaluate the internal abstractions developed through autoregressive action modeling, we compare again to the co-trained baseline, where both metacontroller and base model are jointly optimized through the minimization of \cref{eq:metacontroller-loss}. Consistent with the rate-distortion analysis results (Fig.~\ref{fig:rate_distortion_plot}), the success rate of post-training internal RL remains close to zero. The same holds for CompILE \citep{kipf_compile_2019}, a comparable, previously proposed hierarchical RL method that also relies on variational inference to discover temporally-abstract actions from an unlabeled behavioral dataset. These results again confirm the importance of the initial autoregressive foundation model pretraining phase, followed by base model freezing, for enabling efficient hierarchical RL.

\section*{Discussion}
In this work, we asked whether the latent representations of autoregressive sequence models could be leveraged to develop RL techniques that overcome the inefficiency of token-by-token exploration and reinforcement. We studied this question using tasks that contain multiple subgoals that can be composed together to create the ultimate goal of the task. We first showed that an autoregressive sequence model trained on action and observation sequences from agents trained on simpler versions of the tasks learn representations in their hidden layers that carry information about the subgoals. Next, we demonstrated that these latent representations in the sequence model can be used by a set of internal controllers, provided with the groundtruth subgoals, to solve more complex tasks by compositionally generalizing in time. We then developed a model that uses a metacontroller to select appropriate temporally-abstract actions without receiving the groundtruth subgoal labels. Finally, we showed that directly reinforcing the internal activation controllers generated by the metacontroller enables learning in more complex, hierarchical sparse-reward tasks where other RL techniques fail. Altogether, our results demonstrate that the latent representations of autoregressive sequence models can indeed be leveraged to enable efficient, hierarchical RL.

There is a long-running debate on whether autoregressive next-token predictors can form consistent temporal abstractions and plans \citep{bachmann_pitfalls_2024}, with some researchers dismissing them as ``stochastic parrots'' \citep{bender_dangers_2021}. Our work adds a positive piece of evidence to this question. We chose to study a set of RL environments that fulfill a few key properties we associate with intelligent agents. For an agent to master these environments, it must be able to (i) recombine previous behaviors in novel meaningful ways, (ii) learn from sparse rewards, and (iii) overcome reward sparsity by leveraging imitation learning to infer and repurpose the goal-directed behaviors of other agents. Learning from sparse rewards is arguably the ultimate setting for reinforcement learning, encompassing problem domains ranging from mathematical reasoning and robotic manipulation to scientific discovery in their most ambitious forms. Solving such tasks without reliance on manual reward shaping is a critical step toward autonomous agents capable of navigating complex, open-ended search spaces where the definition of intermediate progress is often unknown.

Despite their simplicity, the environments are challenging enough for standard RL methods to fail, including GRPO (a recent but by now standard method for sparse-reward tasks), as well as CompILE, a previous hierarchical RL algorithm \citep{kipf_compile_2019} that attempts to discover abstract actions from raw unlabeled data, instead of the internal representations of an autoregressive sequence model. The overwhelming success of internal RL over baseline RL algorithms reported here must still be taken with care, however, given the controlled nature of our experimental setup. Investigating and adapting internal RL to larger-scale models and tasks is an important direction of future work.

A number of prior analyses have probed the internal representations of autoregressive models, looking for temporal abstractions and plans. A recent exciting study provided compelling evidence for planning in LLMs asked to write rhyming poems \citep{lindsey_biology_2025}, and earlier probing work found that hidden LLM states have some predictive power over a short number (four) of future tokens \citep{pal_future_2023}. Another line of prior work has focused on models trained from scratch in controlled environments, as we do here, notably in games such as Othello \citep{li_emergent_2023,nanda_emergent_2023} or chess \citep{jenner_evidence_2024,karvonen_emergent_2024}. To the best of our knowledge, we are the first to consider continuous environments with a hidden, discrete, hierarchical task structure. Despite being trained by gradient descent and only employing continuous units (both within the base SSM next-token predictor and the metacontroller) the models nonetheless discovered the underlying discrete latent task structure. In particular, the metacontroller developed sparse, quasi-binary switching units. Moreover, our findings complement recent analyses of convolutional LSTM policies trained by end-to-end RL to play the Sokoban game \citep{bush_interpreting_2025,taufeeque_planning_2024}. These studies showed that RL led to the development of planning subroutines that unfold over multiple timesteps, like the goal-reaching policies that we found within self-supervised autoregressive models. We complement these studies by focusing on autoregressive transformers and SSMs trained on a next-token prediction objective, the current workhorse of artificial intelligence systems.

Schmidhuber theorized in a seminal paper \citep{schmidhuber_learning_2015} that a wake-sleep training loop iterating between training a history compressor through self-supervised learning (SSL), and letting a controller use the internal representations of the former to generate new experiences through RL, would lead to the acquisition of evermore complex capabilities, including the ability to form and exploit temporal abstractions and plans. Here, we provide both a concrete neural architecture following this philosophy, and a set of experimental results backing these claims. Interestingly, we begin to see the benefits of alternating between SSL and RL in large-scale models. For instance, DeepSeek-R1 \citep{guo_deepseek-r1_2025} training also involved one iteration of the RL-SSL cycle, albeit with additional human curation involved in the (post-RL) SSL phase, and with RL still done at (raw) output action level.

Our model also displays similarities to LeCun's joint embedding predictive architecture \citep[JEPA;][]{lecun_path_2022}. In particular, the metacontroller introduced here is similar to the JEPA configurator module, as both are in charge of modulating a general world model and policy in service of a given goal or task. However, JEPA is a proposal for learning abstract observation and action representations without an autoregressive predictive model, whereas next-action prediction is precisely at the center of our approach. In fact, we show that learning a (raw) action predictor is partly what enables discovering how to decompose a task into a sequence of subgoals, one of the open problems in the JEPA proposal.

The overwhelming advantage of internal RL over standard RL finetuning reported in this paper deserves further investigation in real-world environments. A direction that seems particularly worthy of pursuing is LLM reasoning. There is growing interest in reasoning methods that leverage the internal representations of LLMs for reasoning, mainly exploring recurrent iteration in neural activation space \citep[e.g.,][]{hao_training_2024,saunshi_reasoning_2025,shen_codi_2025,geiping_scaling_2025}. The metacontroller model presented in our paper is complementary to these efforts, and may itself benefit from additional recurrence. Instead, the key innovation lies on the discovery of latent variables that compress time dynamically. This has the potential to cut the search space in a reasoning problem and thereby increase RL efficiency, as it did in a dramatic way in the problems considered here. A first step in this direction was taken by Kong et al.~\citep{kong_latent_2025}, who pretrained through variational methods a language model with a stochastic latent variable, and already saw promising results on reasoning benchmarks.

Finally, our results open a new avenue for model interpretability and control at scale. Similarly to sparse autoencoders (SAEs), a popular method for model interpretability and steering, the metacontrollers introduced in this work can be trained through scalable self-supervised learning and employ an encoder-decoder-type architecture. However, the two models otherwise have significant differences. While SAEs are trained on instantaneous internal activation reconstruction, metacontrollers are predictive and interventive, trained to directly lower output next-token prediction error by intervening on the residual stream. Moreover, they maintain internal state, whereas SAEs are instantaneous. Metacontrollers are thus by design likely better suited if the goal is foundation model control, and they offer the possibility of discovering interpretable interventions that run over an extended period of time. We are excited about the prospect of investigating whether these capabilities translate to larger-scale models such as LLMs.

\paragraph{Acknowledgements.} We would like to thank Razvan Pascanu, Jörg Bornschein, Rajai Nasser, Marissa A.~Weis, James Evans, Eric Elmoznino, Sangnie Bhardwaj, Charlotte Frenkel, Anoop Sinha, Zoltan Szabadka, Dileep George, Kevin P.~Murphy and Doina Precup and her lab members for helpful comments and discussions, as well as Yul Kwon and Alice Guan for overall support.

\bibliography{UAP}

\appendix

\renewcommand\thefigure{A\arabic{figure}}
\setcounter{figure}{0}
\renewcommand{\thetable}{A\arabic{table}}
\setcounter{table}{0}

\section{Environment details}\label{app:environmental_details}

\subsection{Gridworld-pinpad}

Our grid world environment, referred to as gridworld-pinpad in the Appendix, is inspired by the previously proposed visual Pin Pad benchmark \citep{NEURIPS2022_a766f56d}. In our version, an agent is located in a grid world, together with uniquely colored cells (also referred to as objects). Within a task, the agent needs to step on a sequence of colored cells in a task-specific order.

\subsubsection{Markov decision process specification}

\begin{itemize}
    \item \textbf{Task:} A task is specified by a sequence of colored cells to visit.
    \item \textbf{State:} The world is a 2D grid of size $G$-by-$G$. There are $O$ unique colored cells placed on the grid, as well as $W$ walls. At any given moment, the agent occupies one of the $G^2-W$ cells that are not wall cells. Finally, the environment state also keeps track of what colored cells the agent has visited so far in the episode.
    \item \textbf{Action:} There are 4 actions corresponding to the 4 cardinal directions.
    \item \textbf{Dynamics:} Given the action and the agent position, the agent moves to the corresponding direction, except when it is moving towards a wall cell or outside of the grid, in which case the action results in a no-op. A colored cell is considered visited when the agent moves onto the cell from a different cell. If the agent successfully visits all colored cells in the right order, or if the agent visits a colored cell that is not the next cell specified by the task, or if the episode lasts longer than $T$ steps, the episode ends. 
    \item \textbf{Initial state:} At the beginning of every episode, the colored cells and walls, as well as the initial agent position are randomly sampled on the grid, ensuring there is no overlap.
    \item \textbf{Observation:} The agent's observation is the one-hot encoding of which object/wall is present in each cell, as well as the one-hot vector corresponding to the position of the agent, resulting in a $G^2(W+O+1)$-dimensional vector.
    \item \textbf{Reward:} The agent gets a reward of $1$ when successfully completing the task, and $0$ otherwise. 
\end{itemize}

\subsubsection{Task specification and hyperparameters}

For both pretraining and post-training tasks, we use $G=7$, $O=8$, $W=4$, and $T=100$.

Numbering the colors from $0$ to $7$, the list of pretraining tasks can be found in Table~\ref{tab:grid_task}. In this setup, the abstract subgoals combined to comprise the compositional final tasks, are given by $0-1$, $2-3$, $4-5$, and $6-7$.

\begin{table}[]
\centering
\begin{tabular}{l}
0-1-4-5-0-1 \\
0-1-4-5-2-3 \\
0-1-6-7-2-3 \\
2-3-0-1-4-5 \\
2-3-6-7-2-3 \\
2-3-6-7-4-5 \\
4-5-0-1-4-5 \\
4-5-0-1-6-7 \\
4-5-2-3-6-7 \\
6-7-2-3-0-1 \\
6-7-2-3-6-7 \\
6-7-4-5-0-1 \\
0-1-6-7-4-5 \\
2-3-0-1-6-7 \\
4-5-2-3-0-1 \\
6-7-4-5-2-3
\end{tabular}
\caption{\textbf{Pretraining tasks for gridworld-pinpad.} Each $c_0-...-c_L$ list entry indicates a task consisting in visiting in order the colors $c_0$, $c_1 \dots c_L$ for some length $L$.} \label{tab:grid_task}
\end{table}

We choose the post-training task to be $0-1-2-3-4-5-6-7-0-1-2-3$.

\subsection{Ant-pinpad}

Ant-pinpad is a continuous control counterpart of the aforementioned gridworld-pinpad. The agent controls the classic MuJoCo ant \citep{schulman_high-dimensional_2015}, with the goal of stepping on a sequence of colored cells in a task-specific order.

\subsubsection{Markov decision process specification}

\begin{itemize}
    \item \textbf{Task:} A task is specified by a sequence of colored cells to visit.
    \item \textbf{State:} The state is a 2D plane, divided into grids. The grid is organized identically to that of the gridworld-pinpad, and also includes colored cells and walls. The state is further augmented by the proprioception state of the ant, as well as the precise coordinate of the center of the ant in the grid. Finally the environment state also keeps track of what colored cells the agent has visited so far in the episode.
    \item \textbf{Action:} The action is an 8-dimensional continuous vector representing the torque applied to the ant's eight joints.
    \item \textbf{Dynamics:} Given the action, the ant moves on the 2D plane as usual. When the center of the ant enters a wall cell or whenever the vertical position of the ant's torso falls outside the valid operational range of $[0.2,\  1.0]$, an episode is instantly terminated. A colored cell is considered visited when the ant enters the cell from a different cell. If the agent successfully visits all colored cells in the right order, or if the agent visits a colored cell that is not the next cell specified by the task, or when the episode lasts longer than $T$ timesteps, the episode ends. 
    \item \textbf{Initial state:} At the beginning of every episode, the colored cells and walls, as well as the initial agent position are randomly sampled on the grid, ensuring there is no overlap. We initialize the agent's full MuJoCo state by first setting the torso's $x,y$ position in the plane to the center of the sampled grid cell. Then we add uniform noise that positions the agent in the simulation anywhere within the boundaries of the initial grid cell. We furthermore sample a random yaw-rotation and turn the agent correspondingly. Finally, the initial angles for all joints and initial velocities are sampled uniformly at random within a small range of $0.1$ units around zero.
    \item \textbf{Observation:} The observation consists of the usual proprioception senses of the ant (to which the symlog function was applied, to ensure no excessively large values occur), concatenated with the global $x,y$ ant coordinate (normalized to be between $-1$ and $1$), as well as the relative position of the various colored cells and walls w.r.t.~the ant, and the local coordinate of the ant within the current cell.
    \item \textbf{Reward:} The agent gets a reward of $1$ when the task is successfully completed, and 0 otherwise. 
\end{itemize}

\subsubsection{Task specification and hyperparameters}

For both pretraining and post-training tasks, we use $G=4$, $O=4$, $W=1$, and $T=500$.

The set of pretraining tasks can be found in Table~\ref{tab:ant_task}.

\begin{table}[]
\centering
\begin{tabular}{l}
0-3-2 \\
1-0-3 \\
2-1-0 \\
3-2-1 \\
0-2-0 \\
0-2-1 \\
0-3-1 \\
1-0-2 \\
1-3-1 \\
1-3-2 \\
2-0-2 \\
2-0-3 \\
2-1-3 \\
3-1-0 \\
3-1-3 \\
3-2-0
\end{tabular}
\caption{\textbf{Pretraining tasks for ant-pinpad.} Each $c_0-...-c_L$ list entry indicates a task consisting in visiting in order the colors $c_0$, $c_1 \dots c_L$ for some length $L$.} \label{tab:ant_task}
\end{table}

We choose the post-training task to be $0-1-2-3$.

\section{Additional experimental results}\label{app:additional_exp}

\subsection{Belief state probing}

\begin{figure}[]
    \centering
    \includegraphics[width=0.6\linewidth]{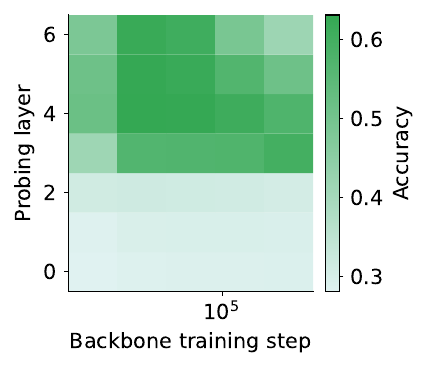}

        \caption{\textbf{Belief state probing at different network layers and base model training steps.} The latent goals governing the behavior of the data forced through a frozen pretrained sequence model become linearly decodable from residual stream activations deep in the network. Here, we display the final accuracy of linear probes trained to predict the latent goals from the residual stream activations of a frozen 6 layer transformer. We vary the training steps of the backbone and the layer at which the probe is plugged into the sequence model and report the mean performance over 10 backbone seeds.}
        \label{appfig:belief_state_probing}
\end{figure}

Fig.~\ref{appfig:belief_state_probing} displays the performance of linear probes predicting latent subgoals from the residual stream activations of a pretrained and subsequently frozen transformer in the gridworld-pinpad environment. These linear probes are obtained by following the procedure detailed in Section~\ref{app:details_belief}. Importantly, these subgoals are not explicitly encoded in the data forced through the sequence model. Nonetheless, throughout training on a large corpus of unannotated goal-directed behavior the sequence model develops internal representations of the subgoals. These internal representations get linearly decodable deep in the network (with the accuracy jumping from 30\% to about 50\% roughly in the middle of the model). Interestingly, close to the output layer of the model (layer 6 in Fig.~\ref{appfig:belief_state_probing}) the performance of linear probes deteriorates when plugged into a backbone trained beyond 100K steps.

\subsection{Effect of sequence model training hyperparameters on the abstract action representations}\label{appsec:hp_scan}

\begin{figure*}[ht]
    \centering
    \hspace{-0.3cm}
    \begin{minipage}{.23\textwidth}
    \includegraphics[width=1\textwidth]{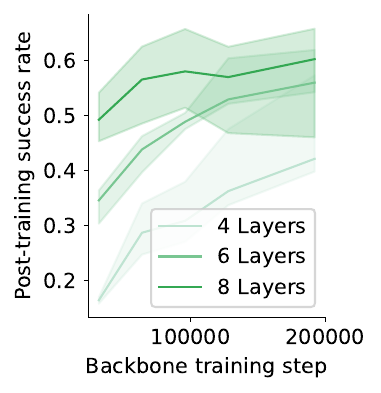}
    \end{minipage}
    \begin{minipage}{.23\textwidth}
    \includegraphics[width=1\textwidth]{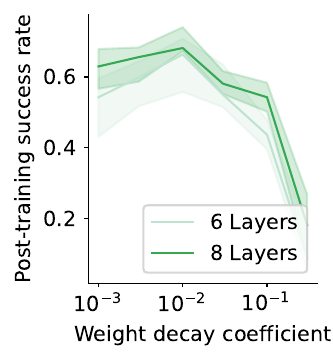}
    \end{minipage}
    \begin{minipage}{.225\textwidth}
    \includegraphics[width=1\textwidth]{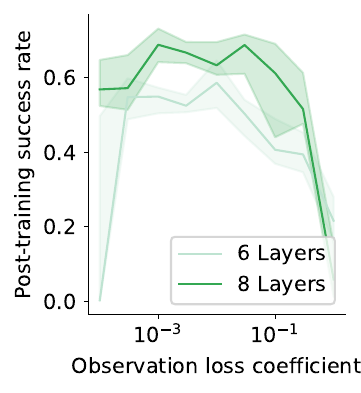}
    \end{minipage}
    \begin{minipage}{.23\textwidth}
    \includegraphics[width=1\textwidth]{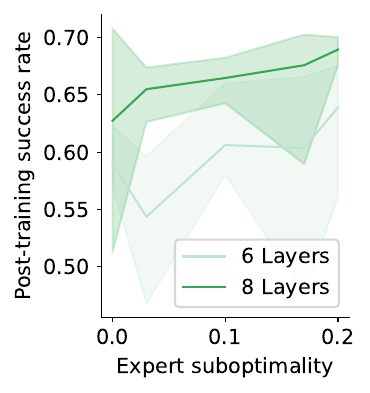}
    \end{minipage}
    \hspace{0.1cm}
    \caption{\textbf{Effect of sequence model training step (left), weight decay (center left), auxiliary observation loss (center right), and expert suboptimality $\epsilon$ (right) during pretraining on the controllers' compositional generalization.} The solid line represents the median performance over 10 runs, 1 seed for each of the 10 pretrained models, and the shaded area indicates the spread between the $25^{\text{th}}$ and $75^{\text{th}}$ quantiles.
    }\label{appfig:wd_obs_scaling}
\end{figure*}

In this section, we investigate the effect of various hyperparameter choices during sequence model training on the internal abstract action representation of the base autoregressive model. For all experiments, we use the gridworld-pinpad environment. We measure the quality of abstract action representation following the procedure outlined in Section~\ref{app:compgen_exp}, and by evaluating the compositional generalization of the obtained controllers on post-training tasks. For all experiments, we use the same hyperparameters as detailed in Section~\ref{app:compgen_exp}, unless specified otherwise. The results are presented in Fig.~\ref{appfig:wd_obs_scaling}.

\paragraph{Sequence model training steps.} For all base autoregressive model depths (4,6 and 8), we notice that  longer sequence model training generally leads to better internal abstract action representation, such that the controllers generalize better to the post-training task set.

\paragraph{Sequence model training weight decay.} For all base autoregressive model depths, we notice that weight decay during sequence model training is beneficial for internal representation. Interestingly, too much weight decay also degrades the representation, which points to a critical regularization trade-off that has been previously reported in foundation models \citep{kobayashi2024weight}.

\paragraph{Observation auxiliary loss.} Next, we observe that some amount of auxiliary loss (i.e.~training to predict the next observation as well as action) is beneficial to building internal abstract action representation. With very low coefficient for the auxiliary loss, we noticed that some models completely failed to learn the representation; however we suspect this behavior is an artifact of our particular environment rather than a general trend. 

\paragraph{Expert suboptimality.} Finally, we investigate the effect of the suboptimality of the demonstrations used during pretraining on the resulting abstract action representation. We achieve this by replacing the expert policy by an $\epsilon$-noisy one, where at every timestep, with probability $\epsilon$, a random (non terminating) action is taken. We see that the abstract action representation is robust against such suboptimality.

\subsection{Unsupervised abstract action discovery}

\subsubsection{Temporal abstraction in the gridworld}
\begin{figure}[]
    \centering
    \hspace*{.15cm}\includegraphics[width=0.48\textwidth]
    {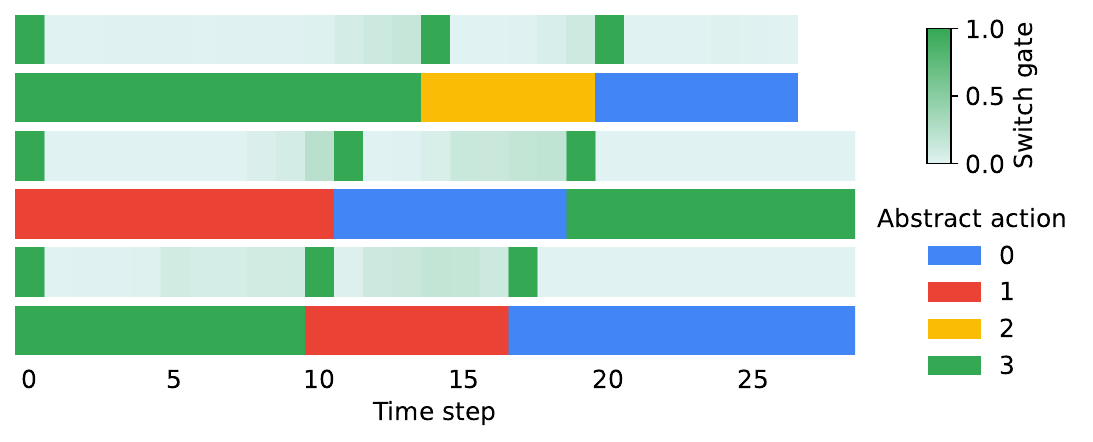}
    \caption{\textbf{Self-supervised metacontroller discovers temporally-abstract actions, gridworld-pinpad.} Three example trajectories from the gridworld-pinpad control environment showing the switch $\beta_t$ used for temporal integration at each timestep, and the groundtruth abstract action being performed (color-coded). Switching ($\beta_t \approx 1$) coincides with a change in the abstract action being performed.}\label{appfig:cossim_grid}
\end{figure}

In Fig.~\ref{appfig:cossim_grid}, we analyze the temporal abstraction discovered in the gridworld-pinpad setting (c.f. Fig.~\ref{fig:switch-gate-analysis} for the respective ant-pinpad results), by plotting the switching gate values $\beta_t$ against groundtruth abstract actions $g_t$. Similarly to the ant-pinpad setting, we find that the metacontroller essentially recovers the groundtruth abstract actions by the switch gate learning to behave in a quasi-binary fashion.

\subsubsection{Quality of abstract actions}\label{app:abstract_action}

Figures~\ref{appfig:cossim_grid} and ~\ref{fig:switch-gate-analysis} reveal that the temporal abstractions discovered by the metacontroller during self-supervised learning reflect the ground truth structure of the underlying task. In particular, the switching unit aligns with compositional abstract subgoals governing the observed data in a quasi-discrete fashion.

In this section, we focus instead on the controller latent code $z$, and provide evidence that the latent space encodes the actual subgoal-seeking abstract actions that constitute the compositional task, in a context-agnostic manner. To achieve this, we focus on the ant-pinpad environment, and follow the following procedure:

\begin{enumerate}
    \item For a handful of grid configurations, we first perform an unconditioned rollout, i.e. a rollout in the environment using the sequence model and the trained metacontroller while sampling the $z$ from the Gaussian prior, instead of the variational distribution.
    \item Next, for each object, we consider unconditioned rollout trajectories that correspond to the agent visiting that object (and nothing else), and collect the latent codes $z$ that were active at the time of visit. We hypothesize these latent codes to encode the subgoal seeking abstract action towards the corresponding object.
    \item Finally, we use those latent codes in different scenarios, and demonstrate that the same latent code's subgoal seeking property generalizes to other situations. 
\end{enumerate}

\begin{figure*}[ht] 
    \centering
    \includegraphics[width=\textwidth]{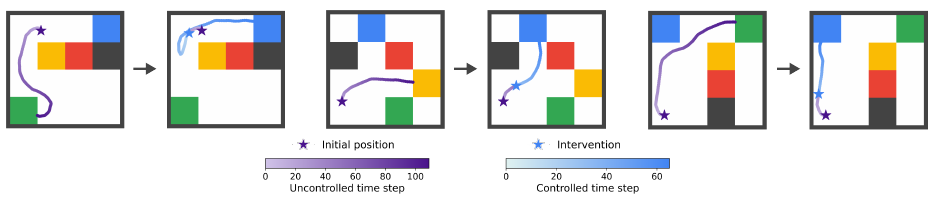}
    
    \caption{\textbf{Controller latent codes implement abstract actions.} Here, we illustrate the effect of a latent controller code implementing the abstract action ``go to blue'' in ant-pinpad, when forcing a switch at an arbitrary time. The three pairs display a trajectory without intervention by the metacontroller (left) vs.~the one with the metacontroller running on a latent code corresponding to ``go to blue'' (right) respectively. The same controller latent code successfully steers the ant towards the desired color in different context, and regardless of the timing at which it is activated. Some trajectories demonstrate backtracking behavior when the control is applied. 
    }
    \label{appfig:examples_abs_action}
\end{figure*}

\paragraph{Generalization to new configurations and switching timing} 
Here, we investigate the ability of the latent code to generalize to new grid configurations and unseen switching times. The metacontroller is trained on successful, nearly-optimal trajectories where agents rarely demonstrate "backtracking" -- behavior where an agent turns away from one object to seek another. Consequently, it is non-trivial whether a latent code injected mid-rollout can override the base model’s current trajectory. As shown in Fig.~\ref{appfig:examples_abs_action}, injecting a ``go to blue'' latent code at timestep 30 causes the agent to immediately correct its course, even if it was previously moving toward a different object. This intervention increases the goal-reaching success rate from $23\%$ in the uncontrolled baseline to $36\%$. This is significant, considering that the latent codes were generated for different configurations and are the result of a noisy sampling.

\paragraph{Generalization to out of distribution sequences} 

We further test whether these codes can force behavior that is explicitly absent from the training data. In the ant-pinpad environment, the agent is never trained to seek object 1 immediately after object 0 (c.f. Section~\ref{app:environmental_details}). By manually activating the latent code for object 1 after the agent reaches object 0, we find the success rate for this OOD transition rises from $10\%$ (baseline) to $24\%$. Note that this also tests whether the same latent codes can generalize to a new position in the sequence, since they were collected from trajectories where the ant visited the corresponding object as the first object.

Ultimately, these results indicate that the metacontroller does not merely learn to segment time, but successfully discovers a compact, steering-capable representation of functional intent—providing the necessary 'options' for internal RL to perform efficient credit assignment in complex, hierarchical tasks.

\subsection{Internal reinforcement learning}
\begin{figure}[htb]
    \centering
    \hspace{-0.3cm}
    \includegraphics[width=.49
    \textwidth]{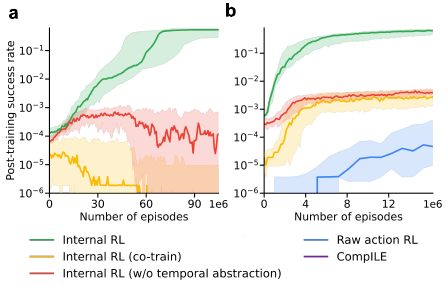}
    \caption{ \textbf{Internal reinforcement learning solves sparse-reward compositional tasks where standard methods fail.} While some baselines see non-zero success rates at some time during RL training they fail to translate these successes into a policy maximizing reward. In turn, internal RL manages to successfully optimize the reward throughout training.
    \label{fig:internal-RL-logscale}}
\end{figure}

We complement the main text Fig.~\ref{fig:internal-RL} by showing the same plot in log scale in Fig.~\ref{fig:internal-RL-logscale}. First, we notice that our internal RL methods achieves the highest success rate at the beginning of training, indicating that noise injection in the residual stream is useful for exploration, compared to  exploration done by sampling raw actions alone. At the same time, despite their high initial success rate, the baselines completely fail at exploiting the experience to reinforce their success. This indicates that proper temporal abstraction is beneficial not only for exploration, but also exploitation as well.

\section{Experimental details}
\label{app:experimental_details}
Here, we describe the details of training the sequence models (cf. Appendix~\ref{app:sequence_model}) as well as several variations of the controller (cf. Appendix~\ref{app:meta_controller}) acting on the then fixed sequence models to produce the results demonstrated in this paper. Moreover, in the final subsection we report the hyperparameters for each experiment.

\subsection{Pretraining of sequence models} \label{app:pretraining_details}
As a prerequisite for the main experiments that all involve tampering with the activations of a pretrained model, the base sequence models $f_{\theta}$ are first trained to autoregressively predict next actions $a_{t}$ and next observations $o_{t+1}$ given a sequence of observations $o_{1:t}$, on a set of meaningful expert trajectories. The details of the expert trajectory generation and sequence model training are given in the following.

\subsubsection{Expert trajectory generation} 
Given an environment and a set of pretraining tasks, expert trajectories are a set of corresponding successful trajectories. The trajectories are not necessarily optimal. 

For the gridworld-pinpad environment, we analytically solve (via dynamic programming) for the stochastic policy that solves the shortest path problem, and, at every timestep, replace the action by a random (non terminating) action with probability $\epsilon$. For all results unless explicitly specified, we chose $\epsilon=0$, but show the robustness against such noise in Section~\ref{appsec:hp_scan}.

For the ant-pinpad environment, we obtain the expert trajectories by training an RL agent. In order to train a single agent for the different task, we augment the observation in the following way: for each cell of the grid, the agent is given an additional 4-dimensional 1-hot vector, indicating one of the 4 cardinal directions the agent must move towards to follow the shortest path at the grid-level. Furthermore, an additional intrinsic reward corresponding to the dot product between the agent's velocity and this direction is given. The agent is trained by PPO \cite{schulman2017proximal}, see Table~\ref{tab:HparamsExpertAnt} for the hyperparameters used.

Note that the resulting expert trajectories are not always successful. The success rate for the pretraining tasks is at 0.8, and for the post-training tasks at 0.7. The success rate in Figure~\ref{fig:internal-RL} is normalized by this score.

\HparamsExpertAnt

\subsubsection{Sequence model training} 
Given a dataset $D$ of expert trajectories, the sequence models are trained to maximize the log-likelihood of the data
\begin{equation}\label{eqn:seq_modeling_loss}
    \begin{aligned}
        &\max_{\theta} \log \prod_{(o_{1:T+1}, a_{1:T}) \sim D} \prod_{t=1}^{T}p_{\theta}(a_{t}|o_{1:t})p_{\theta}(o_{t+1}|o_{1:t})
        \\ = &\max_{\theta} \sum_{(o_{1:T+1}, a_{1:T}) \sim D} \sum_{t=1}^{T} \log p_{\theta}(a_{t}|o_{1:t}) + \log p_{\theta}(o_{t+1}|o_{1:t}).
    \end{aligned}
\end{equation}
Switching the sign and reweighting the observation component with a coefficient $\lambda$ yields the loss function presented in the main text, repeated below for convenience:
\begin{equation}\label{eqn:seq_modeling_objective}
    \begin{aligned}
        \min_{\theta} L(\theta) = \min_{\theta}\sum_{(o_{1:T+1}, a_{1:T}) \sim D} \sum_{t=1}^{T} &-\log p_{\theta}(a_{t}|o_{1:t}) 
         \\
         &- \lambda \log p_{\theta}(o_{t+1}|o_{1:t}).
    \end{aligned}
\end{equation}

Table~\ref{tab:HparamsBackboneGrid} summarizes the hyperparameter choices for training sequence models on the discrete gridworld and Table~\ref{tab:HparamsBackboneAnt} those for the ant. For the sequence models, we use the hyperparameters specified in Table~\ref{tab:HparamsSSM} for SSMs and those specified in Table~\ref{tab:HparamsTF} for transformers. We use SSMs for the ant-pinpad and transformers in the gridworld-pinpad.

\HparamsBackboneGrid
\HparamsBackboneAnt

\subsubsection{Seed}
For both environments, we pretrain 10 such sequence models with different seeds.

\subsection{Belief state probing}\label{app:details_belief}
Given a pretrained sequence model $f_{\theta}$ optimized to maximize Equation~\ref{eqn:seq_modeling_objective}, we train linear probes to predict the latent subgoals governing a sequence at hand. More formally, given a sensori-action sequence $(o_{1:T+1}, a_{1:T})$ we train a linear probe $U_l \in \mathbb{R}^{n_e ,n_g}$ to predict the latent subgoal $g_t$ from the residual activation $e_{t,l}$ at layer $l \in {0,...,L}$ at every timestep $t$. Here, $n_e$ and $n_g$ denote the residual stream dimension and total number of subgoals in the dataset respectively and $L$ is the number of layers of $f_{\theta}$. The belief distribution over the subgoals at timestep $t$ is parameterized as
\begin{align}\label{eqn:belief_state_prob}
    p(g_t=g|e_t) = \mathrm{softmax}(U_le_{t,l})_g.
\end{align}
The parameters $U_l$ are trained to minimize the cross-entropy loss
\begin{equation}\label{eqn:loss_belief}
    \begin{aligned}
        L(U_l)= \sum_{(o_{1:T+1}, g_{1:T}) \sim D} \sum_{t=1}^{T} -\log p(g_t|e_{1:t, l})
    \end{aligned}
\end{equation}
Table~\ref{tab:HparamsBeliefGrid} summarizes the hyperparameter choices for these experiments.

\HparamsBeliefGrid

\subsection{Controller compositional generalization}\label{app:compgen_exp}
Given a pretrained and subsequently fixed sequence model $f_{\theta}$, a modification of the metacontroller $c_{\phi}$ as defined in Appendix~\ref{app:meta_controller} is inserted into the base autoregressive model at some layer $l \in {0,...,L}$. Here, $L$ is the total number of layers of the base autoregressive model. The metacontroller used in these experiments deviates from the vanilla version in how the latent codes $z_t$ are computed. Instead of sampling $\tilde{z}_t$ from a normal distribution and then temporally integrating according to
\begin{equation*}
    z_t = \beta_t \odot \tilde{z}_t + (1-\beta_t)\odot z_{t-1}
\end{equation*}
$\beta_t=1$ is forced for all $t$. Moreover, in these experiments, the ground truth information about abstract behaviour is injected via $\tilde{z}_t$. In particular, the expert trajectories $(o_{1:T+1}, a_{1:T})$ are annotated with the IDs of the abstract actions that governed the behavior during generation. Note that, contrary to the raw action $a_t$, the abstract action and hence the identifier $\mathrm{ID}_t$ provided to the metacontroller only change sparsely in time. Formally, given the ground truth labels, the controller latent code at time step $t$ is given by 
\begin{equation*}
    z_t = \tilde{z}_t = \mathrm{onehot}(\mathrm{ID_t}, K),
\end{equation*}
the onehot encoding of $\mathrm{ID}_t$. Here, $K$ denotes the total number of unique abstract actions in the dataset $\mathcal{D}$.
With access to this additional privileged information, we train the parameters $\phi$ to further maximize the data log-likelihood (cf. Equation~\ref{eqn:seq_modeling_objective}).

Table~\ref{tab:HparamsControllerCompGrid} summarizes the hyperparameter choices for the controller compositional generalization on the discrete gridworld and Table~\ref{tab:HparamsControllerCompAnt} those for the ant.

\HparamsControllerCompGrid
\HparamsControllerCompAnt

\subsection{Unsupervised abstract action discovery}\label{app:metacontroller_training}
Given a pretrained, frozen base sequence model $f_{\theta}$, the metacontroller $c_{\phi}$ as described in Appendix~\ref{app:meta_controller} is inserted into the base model at some layer $l \in {0,...,L}$ where $L$ is the base model depth. With $\theta$ frozen, the metacontroller parameters $\phi$ are trained to (further) minimize a regularized NLL. Note that the metacontroller $c_{\phi}$ learns to generate and make use of an acausal embedding $s(e_{1:T})$. Thus, by controlling the base model, the metacontroller can minimize the NLL beyond the loss-level attained by the causal base autoregressive model. Beyond optimizing Equation~\ref{eqn:seq_modeling_objective}, the posterior over controller latent codes

\begin{equation*}
    \tilde{z}_t \sim \mathcal{N}(z_\text{enc} ; \mu_t, \Sigma_t)
\end{equation*}
is regularized so that at test time meaningful controller latent codes can be sampled from the prior $\mathcal{N}(0, I)$. To allow this, the Kullback-Leibler divergence between both distributions 
\begin{equation}
    D_{\mathrm{KL}}\big(\mathcal{N}(\mu_t, \Sigma_t) \,\|\, \mathcal{N}(0, I)\big)
\end{equation}
is added to the NLL objective. Putting everything together, and adding regularization strength $\alpha$ the metacontroller $c_{\phi}$ is trained to minimize the loss
\begin{equation}\label{eqn:ssl_skill_discovery_objective}
    \begin{aligned}
    L(\phi) = \sum_{(o_{1:T+1}, a_{1:T}) \sim D} \sum_{t=1}^{T} &-\log p_{\phi, \theta}(a_{t}|o_{1:t})
    \\ &- \lambda \log p_{\phi, \theta}(o_{t+1}|o_{1:t})
    \\ &+ \alpha D_{\mathrm{KL}}\big(\mathcal{N}(\mu_t, \Sigma_t) \,\|\, \mathcal{N}(0, I)\big)
\end{aligned}
\end{equation}
where $p_{\phi, \theta}$ denotes the probability computed by the sequence model $f_{\theta}$ when controlled by $c_{\phi}$. This objective is motivated as the evidence lower bound (ELBO) in Section~\ref{app:elbo}. Again, note that only the parameters $\phi$ are trained while the sequence model $\theta$ remains frozen.

The hyperparameters for training the metacontroller in gridworld- and ant-pinpad are summarized in Table~\ref{tab:HparamsControllerUnsupGrid} and Table~\ref{tab:HparamsControllerUnsupAnt}, respectively.

\HparamsControllerUnsupGrid
\HparamsControllerUnsupAnt

\subsubsection{Baseline -- forced resets}
This baseline aims to answer the question whether a metacontroller not factorizing the controller latent code $z_t$ into explicit subsequences via $\beta_t$ discovers abstract actions suitable for subsequent internal RL. To do so, we perform the exact same experiment as described so far in this subsection with the only difference that $\beta_t=1$ is forced at every timestep. Hence, $z_t$ is equal to the latent controller code proposal $\tilde{z}_t$.

\subsubsection{Baseline -- metacontroller cotraining}
This baseline investigates whether the two stage approach of first training the sequence model $f_{\theta}$, freezing it, and only then training the metacontroller $c_{\phi}$ yields different results than cotraining both $\theta$ and $\phi$. In this pursuit, we perform 2 experiments: for gridworld, no pretrained $\theta$ is assumed and instead both $\theta$ and $\phi$ are both randomly initialized and jointly trained to optimize the regularized NLL defined in Equation \ref{eq:metacontroller-loss} (else used for training $\phi$ in a frozen $f_{\theta}$). For ant-pinpad, $\theta$ is initialized to the pretrained parameter, but we still jointly train $\phi$ and $\theta$ to optimize the objective. 

\subsubsection{Baseline -- CompILE}

We adapted CompILE \citep{kipf_compile_2019,jiang_learning_2022} as best as possible to our setting. 

On a high level, CompILE is very similar to our cotraining baselines: it is a latent variable model (albeit with a different set of latent variables) which takes a sequence of observations and output, for each timestep, a continuous latent variable $z$ drawn from a Gaussian that then condition a policy trained to imitate the action in the trajectory. Similarly to us, it is a variational inference approach to discovering the abstract actions, except that it does not leverage the internal representation of a pretrained model. CompILE also infers the switching latent variables $\beta$, and requires a prior distribution over the switching rate and the maximum number $M$ of abstract actions (or segments) in all sequences. 

To make things comparable, we adopt CompILE to our architecture by drop-in replacing the metacontroller by the CompILE module which generates the latent code $z$, while keeping everything else identical. In particular, the same sequence model architecture is used and internally controlled by $z$. Nevertheless, to remain close to the original architecture of CompILE, the module generating the latent code takes as input the raw input, instead of the residual stream activation. To compensate for the potential loss in expressivity, we use as the recurrent encoder of the module the same architecture as the first half of the sequence model. The parameters for the distribution of the latent variables are then generated by a one hidden layer MLPs with hidden dimension $n_h$. Due to the difficulty of performing parallel inference when using the algorithm, the sequence model parameters $\theta$ are initialized to the pretraining value, such that training can be shortened.

Table~\ref{tab:HparamsControllerUnsupCompileGrid} (resp. ~\ref{tab:HparamsControllerUnsupCompileAnt}) shows the hyperparameter used for gridworld (resp.~ant-pinpad).

\HparamsControllerUnsupCompileGrid
\HparamsControllerUnsupCompileAnt

\subsubsection{Metacontroller training dataset} \label{app:metacontroller_demo}

While our experiments reveal that the emergence of abstract actions in the sequence model is robust to suboptimality (c.f. Appendix \ref{appsec:hp_scan}) the opposite holds for discovering these abstract actions in the frozen sequence model with the metacontroller. Empirically, we observe that as the demonstrations used for training the metacontroller get cleaner (i.e. the closer they resemble the optimal stochastic policy), the ability of the metacontroller to compress the abstract actions onto a latent space improves. Therefore, for training the metacontroller, for ant-pinpad, we generate trajectories by taking the mean of the Gaussian distribution generated by the expert for each timestep instead of sampling from it. Similarly, for grid we set the suboptimality degree $\epsilon=0$, to obtain clean expert demonstration.

\subsubsection{Seed}
For all environments and methods, for each of the 10 pretrained sequence models and each of the hyperparameter configurations, we perform this unsupervised abstract action discovery over 3 different seeds.

\subsection{RL experiments}

\subsubsection{Internal RL}\label{app:internal_rl}

We present in algorithm~\ref{alg:internal_rl_init} the initialization function for the internal RL environment, and ~\ref{alg:internal_rl_step} the effective internal RL environment step function, given the original environment, a pretrained base autoregressive model and corresponding meta controller. Algorithm ~\ref{alg:internal_rl_full} shows the full training loop.

The hyperparameters for training the agent in gridworld- and ant-pinpad are summarized in Table~\ref{tab:HparamsInternalRLGrid} and Table~\ref{tab:HparamsInternalRLAnt}, respectively.

\HparamsInternalRLGrid
\HparamsInternalRLAnt
\begin{algorithm}
\caption{The effective internal RL environment step function}
\label{alg:internal_rl_step}
\SetKwInOut{Input}{require} 
\SetKwFunction{FStep}{step}
\DontPrintSemicolon 

\Input{Original environment $E$, switching unit $f_\text{switch}$, controller decoder $f_\text{hyp}$, model blocks $f_{\text{block}_{:l}}$ up to layer $l$, model blocks $f_{\text{block}_{l:}}$ from layer $l$. The function takes the abstract action $z$, and the internal state $s$ as inputs.}

\BlankLine
\FStep($z, s$):\;
\Indp
$\beta \leftarrow 0$\;
$\text{done} \leftarrow \text{False}$\;
$r_\text{acc} \leftarrow 0$\;
$(e, h_{\text{switch}}, h_{\text{block}_{l:}},h_{\text{block}_{:l}}) \leftarrow s$\;
\While{$\beta < \beta_\text{threshold}$}{
    $U \leftarrow f_\text{hyp}(z)$\;
    $a, h_{\text{block}_{l:}} \sim f_{\text{block}_{l:}}(e + U e, h_{\text{block}_{l:}})$\;
    $o, r, \text{done} \sim E.\text{step}(a)$\;
    $e, h_{\text{block}_{:l}} \leftarrow f_{\text{block}_{:l}}(o, h_{\text{block}_{:l}})$\;
    $\beta, h_{\text{switch}} \leftarrow f_\text{switch}(e, z, h_{\text{switch}})$\;
    $r_\text{acc} \leftarrow r_\text{acc} + r$\;
}
$s \leftarrow  (e, h_{\text{switch}}, h_{\text{block}_{l:}},h_{\text{block}_{:l}}) $\;
\KwRet{$(e, r_\text{acc}, \text{done}), s$} 
\end{algorithm}

\begin{algorithm}
\caption{The internal RL initialization function}
\label{alg:internal_rl_init}
\SetKwInOut{Input}{require} 
\SetKwFunction{FInit}{init} 
\DontPrintSemicolon 

\Input{Original environment $E$, switching unit $f_\text{switch}$, controller decoder $f_\text{hyp}$, model blocks $f_{\text{block}_{:l}}$ up to layer $l$, model blocks $f_{\text{block}_{l:}}$ from layer $l$.}

\BlankLine
\FInit():\;
\Indp 
$o, r, \text{done} \sim E.\text{init}()$\;
$h_{\text{block}_{l:}} \leftarrow f_{\text{block}_{l:}}.\text{init}()$\;
$h_{\text{block}_{:l}} \leftarrow f_{\text{block}_{:l}}.\text{init}()$\;
$h_{\text{switch}} \leftarrow f_\text{switch}.\text{init}()$\;

$e, h_{\text{block}_{:l}} \leftarrow f_{\text{block}_{:l}}(o, h_{\text{block}_{:l}})$\;
$\beta, h_{\text{switch}} \leftarrow f_\text{switch}(e, z, h_{\text{switch}})$\;
$s \leftarrow  (e, h_{\text{switch}}, h_{\text{block}_{l:}},h_{\text{block}_{:l}}) $\;
\KwRet{$(e, r, \text{done}), s$} 
\end{algorithm}

\begin{algorithm}
\caption{The internal RL full algorithm}
\label{alg:internal_rl_full}
\SetKwInOut{Input}{require} 
\DontPrintSemicolon 

\Input{Policy $\pi_\theta$}

\For{epoch $e=1\dots E$}{
$\mathcal{B} \leftarrow []$ \;
\For{batch element $b=1\dots B$}{
$(e,r,done), s \leftarrow \texttt{init}()$ \;
$h_\pi \leftarrow \pi_\theta.\text{init}()$ \;
$\tau \leftarrow []$ \;
\While{not done}{
\# acting on a temporally abstract timescale (see algorithm ~\ref{alg:internal_rl_step}) \;
$z, h_\pi \sim \pi_\theta(e, h_\pi)$ \;
$\tau.\text{append}((e,r,done, z))$ \;
$(e,r,done), s \leftarrow \texttt{step}(z, s)$ \;
}
$\tau.\text{append}((e,r,done, None))$ \;
$\mathcal{B}.\text{append}(\tau)$ \;
}
Update policy $\pi_\theta$ using $\mathcal{B}$ by maximizing the objective in Eq~\ref{appeq:grpo}\;
}
Output $\pi_\theta$ \;
\end{algorithm}

\subsubsection{RL algorithm details}\label{app:grpo}

For all RL experiments, we used an RL algorithm suitable for sparse, single final reward setting. The algorithm is related to the GRPO algorithm, except for the notion of group which is absent in our setting. Similarly to GRPO, we modify the standard Proximal Policy Optimization \citep[PPO;][]{schulman2017proximal} framework by replacing the learned value function (critic) with an empirical advantage estimation.

\paragraph{Objective function.}

We optimize the policy by maximizing a clipped surrogate objective similar to PPO. The loss is defined as:

\begin{align}\label{appeq:grpo}
\mathbb{E}_\tau \Big[ \sum_{t} \min( \frac{\pi_\theta(a_t | s_{1:t})}{\pi_{\theta_{\text{old}}}(a_t | s_{1:t})}, \text{clip}( \frac{\pi_\theta(a_t | s_{1:t})}{\pi_{\theta_{\text{old}}}(a_t | s_{1:t})}, 1-\epsilon, 1+\epsilon)) \mathcal{A}_\tau \Big]
\end{align}
where $\pi_\theta$ is the current policy and $\pi_{\theta_\text{old}}$ is the previous policy, $\mathcal{A}_t$ is the relative advantage of the trajectory $\tau$.

\paragraph{Relative advantage estimation.}

We adopt the critic-free approach to estimating the advantage.

The relative advantage $\mathcal{A}_\tau$ measures how much better (or worse) a specific trajectory $\tau$ is compared to the average quality of the entire batch of size $B$. It is calculated by normalizing the reward $R(\tau)$ relative to the batch's mean $\bar{R}$ and standard deviation $\sigma_R$:
\begin{align*}
&\bar{R} = \frac{1}{B} \sum_{i=1}^{B} R(\tau_i) \\
&\sigma_R = \sqrt{\frac{1}{B} \sum_{i=1}^{B} (R(\tau_i) - \bar{R})^2} \\
&\mathcal{A}_\tau = \frac{R(\tau) - \bar{R}}{\sigma_R + \delta},
\end{align*}
where $\delta$ is a small constant (e.g., $10^{-3}$) to ensure numerical stability and prevent division by zero.

\subsubsection{Baseline -- raw action RL}

For the raw action RL baseline, we simply use the pretrained sequence model without any metacontroller, and finetune it on the post-training task using the RL algorithms described in section~\ref{app:grpo}, in raw action space. Since for the raw action RL baseline, there is no unsupervised abstract action discovery phase, we instead do the RL over 3 seeds. 

\subsubsection{Baseline -- others}
For all other baselines, we simply perform internal RL with the respective metacontrollers obtained during the unsupervised abstract action discovery, cf section ~\ref{app:metacontroller_training}. 

\subsubsection{Seed and Hyperparameter selection}
For each method, we scan over different learning rates $(0.000003,0.00001,0.00003,0.0001,0.0003)$ and pick the learning rate and hyperparameter configuration from the unsupervised abstract action discovery with the best median RL performance over the 10 pretrained model and 3 seed. 

\subsection{Rate-distortion curve}\label{app:rate_distortion}

The rate distortion curve is plotted after performing the unsupervised abstract action discovery with the same hyperparameters as described in section ~\ref{app:metacontroller_training}, with the exception of using the Gumbel-Sigmoid trick for the switching units as derived in~\ref{app:elbo} for  gridworld-pinpad, as it resulted in a cleaner Pareto frontier. As the pretrained sequence models have each different Pareto frontier which would hide its structure, we pick one sequence model at random, and instead do the unsupervised abstract action training with 10 seeds on each of the different KL strength $\alpha$. We do this for our method, as well as the metacontroller cotraining baseline. 

The sum of switching value is computed by hard-thresholding the continuous switching value with $\beta_\text{threshold}=0.5$ as we do in the internal RL (cf algorithm ~\ref{alg:internal_rl_step}), and taking the average sum over trajectories. We manually checked the forget patterns to label whether the switching patterns aligned with the subgoal change. 

\section{Architecture details}\label{app:architecture}

\subsection{Sequence model}\label{app:sequence_model}

We parametrize the base model as an autoregressively-trained multi-layer sequence model $f_{\theta}$. The specific instantiations of $f_{\theta}$ detailed below utilize either standard transformer \cite{vaswani_attention_2017} or recurrent neural network (also commonly referred to as state-space model, SSM) layers. From the latter family (\cite{yang2025parallelizinglineartransformersdelta, yang2024gatedlinearattentiontransformers, yang2025gateddeltanetworksimproving, beck2024xlstmextendedlongshortterm, de_griffin_2024, gu2024mambalineartimesequencemodeling, peng2025rwkv7gooseexpressivedynamic, vonoswald2025mesanetsequencemodelinglocally} and others), we choose the Hawk \cite{de_griffin_2024} due to its simplicity and computational efficiency.

\subsubsection{SSM}\label{app:ssm}

For SSM-based sequence models, we employ a standard pre-normalization layer architecture. Inputs are normalized before being fed into the recurrent Hawk sequence mixing block \cite{de_griffin_2024}, whose output is added back to the residual stream. This is followed by an MLP channel-mixing block that similarly applies normalization to its input before adding its output back to the residual stream.

\HparamsSSM

\subsubsection{Transformer}
For transformer-based models, we employ a standard pre-normalization layer architecture. We first compute relative position embeddings to serve as attention biases. Inputs are then normalized and fed into the Multi-Head Attention sequence mixing block (incorporating these biases), whose output is added back to the residual stream. This is followed by an MLP channel-mixing block that applies normalization to its input before adding its output back to the residual stream.

\HparamsTF

\subsection{Metacontroller architecture}\label{app:meta_controller}
\paragraph{Design principles.} The metacontroller is designed to act inside a frozen, autoregressive sequence model backbone. It does so by modulating the residual stream activations at some backbone layer via simple, internal controllers. Manipulating the residual stream allows the metacontroller to implement temporally abstract actions that turn the sequence model into a subgoal-optimizing policy pursuing a selected goal over multiple raw action timesteps.

These temporally abstract actions implement the subgoals governing the behaviour of the agents whose trajectories constitute the offline data available for metacontroller training. To discover these abstract actions, the metacontroller tracks a recurrent latent variable $z_t$ capturing the subgoal active at step $t$ and then translates it into an action (linear controller). The true posterior $p(z_t|e_{1:T})$ over this latent $z_t$ is inherently acausal since (sub)goals only materialize over an entire trajectory. To make this tangible consider an agent that is placed in the gridworld with the intent to ``go to red''. As the agent takes its first goal-directed actions, an outside observer will have a hard time determining the underlying goal since reaching other colored cells like green might require taking the very same first actions. Only as the action sequence further unfolds and the evidence of the agent's intent becomes conclusive the goal can be identified. These considerations underline that to correctly infer the subgoal at step $t$, in general, the metacontroller needs access to sequence-level information. They also reveal that the purely causal backbone is limited to, at best, discover a powerful online inference algorithm for latent subgoals.

Now, as is the case for real world data, assume that the offline trajectories include the behavior of agents completing subgoals and subsequently switching to new subgoals. It is desirable for the metacontroller to infer the latent $z_t$ in a way that makes the factorization into subgoals accessible. This boils down to parameterizing $z_t$ as a temporal composition of latent codes $\tilde{z}_t$ orchestrated by a switching unit $\beta_t \in [0,1]$. Selecting $\beta_t \approx 1$ implements switching subgoals and, equally important to achieve temporally consistent behaviour, $\beta_t \approx 0$ allows to maintain the previous subgoal. While, as discussed above, the $\tilde{z}_t$ needs to be acausal, $\beta_t$ is parameterized to be causal. This allows the metacontroller to identify when to switch goals at test time (when no acausal information is available). For more justification of this choice, see Section~\ref{app:elbo}.

\begin{figure}[htbp!]
    \centering
     \includegraphics[width=0.95\linewidth]{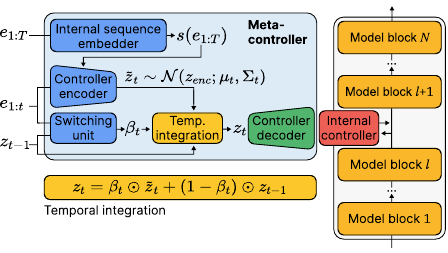}

        \caption{\textbf{Metacontroller architecture} (same as Fig.~\ref{fig:metacontroller-arch}).}
        \label{fig:controller}
\end{figure}

\paragraph{High-level description.} On a high level, the metacontroller $c_{\phi}$ (cf.~Fig.~\ref{fig:controller}) can be viewed as a recurrent hypernetwork \citep{ha_hypernetworks_2017}. It acts inside a frozen sequence model backbone $f_{\theta}$ by emitting \textit{internal, linear controllers} $U_t$ altering the residual activations at timestep $t$. Architecturally, it is an encoder-decoder generative model that allows to sample controllers that, after training, implement abstract actions. First, at every timestep the recurrent \textit{controller encoder} stochastically proposes controller latent codes $\tilde{z}_t$ conditioned on an acausal embedding $s(e_{1:T})$ generated by the \textit{internal sequence embedder}. Also per timestep, the \textit{switching unit} emits a temporal integration rate $\beta_t$. Subsequently, the \textit{temporal integration unit} takes per timestep latent proposals and composes them sparsely in time by applying the temporal integration rate. The temporally integrated latent controller codes $z_t$ are then mapped to instantaneous controllers by the \textit{controller decoder}.

\paragraph{Architectural details.} 
Adopting the standard design principles for sequence models, the pretrained sequence model $f_{\theta}$ is built from $L$ stacked layers (cf. Appendix~\ref{app:sequence_model}). While processing the sequence of inputs $o_{1:T}$ to predict $(a_{1:T}, o_{2:T+1})$ the sequence model generates a sequence of residual stream activations $e_{1:T,l}$ at each layer $l \in \{0,...,L\}$ ($0$ refers to the residual activations immediately before the first layer). The metacontroller operates inside the sequence model at layer $l$ by reading these residual activations $e_{1:T,l}$ and applying an internal, linear controller 
\begin{equation}
    \hat{e}_{t,l} = e_{t,l} + U_{t}e_{t,l}.
\end{equation}
This controlled residual activation $\hat{e}_{t,l}$ is passed to the subsequent model blocks, which are unaware of the intervention. The following details the metacontroller architecture used to learn these interventions from offline data.

The metacontroller keeps track of two summarizing states--a summary state $h_t$ of the history of activations $e_{1:t, l}$ and an embedding $s(e_{1:T,l})$ summarizing the entire sequence of activations $e_{1:T}$. The history is generated by a GRU \cite{cho_properties_2014} and compresses information from past residual activations in its $n_h$ dimensional hidden state
\begin{equation}
    h_t = \text{GRU}(e_t, h_{t-1}).
\end{equation}
This hidden state allows the metacontroller to remember relevant information about the history at test-time. Beyond the history, as discussed above, the metacontroller needs access to sequence level (acausal) information to fulfill its overarching purpose to approximate the posterior $q(z_t | e_{1:T})$. This information is provided through the \textbf{internal sequence embedder} $f_\text{emb}$. It takes the trajectory of residual stream activations $e_{1:T,l}$ produced at layer $l$ and summarizes it in an internal sequence embedding
\begin{equation}
    s(e_{1:T,l}) = f_\text{emb}(e_{1:T,l})
\end{equation}
of dimension $n_{s}$. The internal sequence embedder is parameterized as a SSM (cf. Appendix \ref{app:ssm}) with $L_{emb}$ layers.

Conditioned on this sequence embedding the \textbf{latent proposal mechanism} estimates a distribution over controller latent codes $z_t$ and samples it to produce a $n_z$ dimensional controller latent code proposal $\tilde{z}_t$ at every timestep $t$. The distribution is set to be a normal 
\begin{equation}
    \tilde{z}_t \sim \mathcal{N}(\mu_t, \Sigma_t),
\end{equation}
where $\Sigma_t$ is chosen to be diagonal for computational efficiency. The parameters for mean and variance are produced by the \textbf{controller encoder}
\begin{equation}
    \mu_t, \Sigma_t = f_\text{enc}\left(e_{t,l}, h_{t-1}, s\left(e_{1:T,l}\right)\right).
\end{equation}

Crucially, if this was the final parameterization of the approximate posterior $p(z_t | e_{1:T,l})$ it would not provide a handle on the factorization of the subgoals the agent composed in time when generating its behaviour. As a first step to obtain such a factorization, the metacontroller implements a \textbf{switching unit} producing the temporal integration rate
\begin{equation}
    \beta_t = f_\text{switch}(e_{t,l}, h_{t-1}, z_{t-1}) \in [0,1].
\end{equation}

This integration rate is passed to the \textbf{temporal integration unit} which uses it to combine the latent code proposals $\tilde{z}_t$ sparsely in time. In particular, given $\beta_t$, $\hat{z}_t$, and $z_{t-1}$, the updated latent code is given by the convex combination
\begin{equation}
    z_t = \beta_t \odot \tilde{z}_t + (1-\beta_t)\odot z_{t-1}.
\end{equation}
Observe, that since the $\tilde{z}_t$ are stochastically generated so is $z_t$. 
Moreover, the $\beta_t$ which only rely on causal information and hence can be generated at test time provide a direct handle on the subgoals. When $\beta_t \approx 1$ a new subgoal $\tilde{z}_t $ takes over while $\beta_t \approx 0$ indicates that the previous subgoal $z_{t-1}$ remains a valid explanation for the intent of the agent. This latent controller code $z_t$ is then sent through the \textbf{controller decoder}. The controller decoder is a hypernetwork emitting the \textit{internal controller}
\begin{equation}
    U_t = f_\text{hyp}(z_t).
\end{equation}
As detailed above this linear controller is applied to the residual stream to control the backbone thereby impacting the predicted data log-likelihood computed at the output of the sequence model. Crucially, the described mechanism allows the meta controller to act on extended timescales by maintaining the latent code $z_k$ computed at timestep $k$ for some $n$ timesteps (by setting $\beta_{k+1:k+n-1}=0$). Thereby, since the computation of the hypernetwork is deterministic, the same instantaneous controller $U_k$ can be applied for $n$ timesteps and corresponds to a temporally abstract action.

\subsection{Internal RL policy architecture}\label{app:rl_policy}

Since the residual activation $e_t$ for a single layer does not necessarily contain all information about the raw input history, we use a recurrent policy. A simple 1-layer SSM as described in section~\ref{app:ssm} is used. See Table~\ref{tab:HparamsInternalRLGrid} and ~\ref{tab:HparamsInternalRLAnt} for more details on the architecture.

\section{Additional discussions}

\subsection{Graphical model and ELBO derivation}\label{app:elbo}

Here, we present the graphical model used to derive our unsupervised objective. We denote by $e_t$ the residual stream activation at time $t$, $a_t$ the action, $z_t$ the abstract action of which $a_t$ is part, and $\beta_t$ the random variable indicating a change in the abstract action, i.e., $z_{t-1} \neq z_t$ if $\beta_t=1$. The assumed generative model is as follows:
\begin{align*}
    &p(\beta_{1:T}, z_{1:T}, a_{1:T} \mid e_{1:T}) \\
    = &\prod_t p(\beta_t \mid e_{1:t}) p(z_t \mid z_{t-1}, \beta_t) p(a_t \mid z_t, e_{1:t})
\end{align*}
where $ p(z_t \mid z_{t-1}, \beta_t) = \mathbb{1}_{z_{t}=z_{t-1}}$ if $\beta_t=0$ else $\mathcal{N}(z_t \mid 0,I)$.

We want to optimize the likelihood of observing the sequence of actions by maximizing the following evidence lower bound (ELBO):
\begin{align*}
    &\log p(a_{1:T} \mid e_{1:T})\\
    \geq & \int_{\beta, z} q(\beta_{1:T}, z_{1:T} \mid e_{1:T}, a_{1:T}) \frac{p(\beta_{1:T}, z_{1:T}, a_{1:T} \mid e_{1:T})}{q(\beta_{1:T}, z_{1:T} \mid e_{1:T}, a_{1:T})}
\end{align*}
where $q$ is the variational distribution.
This lower bound holds for any choice of $q$. Following the graphical model, $q$ can be factorized as follows:
\begin{align*}
    &q(\beta_{1:T}, z_{1:T} \mid e_{1:T}, a_{1:T}) \\
    = &\prod_t q(\beta_t \mid e_{1:t}, a_{1:T}) q(z_t \mid z_{t-1}, \beta_t, a_{1:T})
\end{align*}where $ q(z_t \mid z_{t-1}, \beta_t,  a_{1:T}) = \mathbb{1}_{z_{t}=z_{t-1}}$ if $\beta_t=0$ else  $\mathcal{N}(z_t \mid  \mu_t(a_{1:T}), \Sigma_t(a_{1:T}))$ where $\Sigma$ is diagonal.

The ELBO can then be written as 
\begin{align*}
    \text{ELBO}=& \sum_t \log p_\phi(a_t \mid z_t, e_{1:t}) \\
    &+ D_\text{KL}(q(\beta_t \mid e_{1:t}, a_{1:T}) || p(\beta_t \mid e_{1:t}))\\
    &+ D_\text{KL}(q(z_t \mid z_{t-1}, \beta_t, a_{1:T}) || p(z_t \mid z_{t-1}, \beta_t)).
\end{align*}
It can be shown that the last term can be further decomposed as 
\begin{align*}
&D_\text{KL}(q(z_t \mid z_{t-1}, \beta_t, a_{1:T}) || p(z_t \mid z_{t-1}, \beta_t)) \\
=&
\begin{cases}
0 & \text{if } \beta_t = 0 \\
D_\text{KL}(\mathcal{N}(z_t \mid \mu_t(a_{1:T})  ,\Sigma_t(a_{1:T}) )  ||\mathcal{N}(0,I)) & \text{if } \beta_t = 1.\\
\end{cases}
\end{align*}

\paragraph{Continuous relaxation.} In order to improve stability during training, we make a continuous relaxation of the latent variable $\beta$ sampled. In principle, this can be done with the Gumbel-Sigmoid trick, but in our experiments we simply used the probability as the latent variable.

We modify the prior and variational distribution on $z$ to be the continuous relaxation, i.e.,
\begin{align*}
     &p(z_t \mid z_{t-1}, \beta_t)=\mathcal{N}(z_t \mid (1-\beta_t)z_{t-1}, \beta_t^2 I)\\
     &q(z_t \mid z_{t-1}, \beta_t,  a_{1:T})\\
     &=\mathcal{N}(z_t \mid \beta_t \mu_t(a_{1:T}) + (1-\beta_t)z_{t-1} ,\beta_t^2\Sigma_t(a_{1:T})).
\end{align*}
This recovers the previous behavior when $\beta_t$ equals $0$ or $1$.

In the continuous case, it can be shown that the KL divergence is 
\begin{align*}
&D_\text{KL}(q(z_t \mid z_{t-1}, \beta_t, a_{1:T}) || p(z_t \mid z_{t-1}, \beta_t)) \\
=&D_\text{KL}(\mathcal{N}(  \mu_t(a_{1:T})  ,\Sigma_t(a_{1:T}) ) || \mathcal{N}( (0,I) ).
\end{align*}

\paragraph{Further assumptions.} In our experiments, we further modify the variational distribution on $\beta$ such that $q(\beta_t \mid e_{1:t}, a_{1:T}) \approx q(\beta_t \mid e_{1:t})$, i.e., we drop the conditioning on the future. This is done such that during internal RL, the switching signal can be emitted causally, and eliminates the prior matching term for the switching module. This assumption was made in our experiments since we assumed the residual activation to be highly informative of when switches should occur, in the environments considered. In general however, we can relax this assumption by keeping the future conditioning, but distilling the switches to an unconditioned module.

\subsection{Internal RL vs RL with reparametrization trick}
\label{app:internal_rl_vs_reparam}
As explained in Section~\ref{app:internal_rl}, internal RL learns a policy over the discovered abstract action space of $z$ by treating the rest of the architecture as part of the environment, and applying reinforcement learning directly to $z$, with temporal abstraction. However, there are other ways to use the discovered abstract actions than the proposed internal RL.  One perhaps more straightforward way to use the metacontroller, is to treat this policy as a noise-injecting submodule of the overall architecture which is still trained by reinforcement learning in raw action space, by backpropagating through the base autoregressive model, to the policy using e.g. the reparametrization trick. In this section, we analytically contrast these 2 options, discuss their respective advantages, and motivate why we believe the internal RL is interesting in general.

To simplify the analyses, we make a few assumptions:
\begin{itemize}
    \item We remain in the outcome-supervision setting: a single reward $r_T$ is provided at the last time step $T$.
    \item The switching happens $M$ times, at $(t_m)_{1\leq m\leq M}$.
    \item The abstract action policy has a fixed variance, i.e., it outputs $z_t=\mu(s_t) + \epsilon_t$ where $s_t$ is the history of observations up to $t$, $\epsilon_t \sim \mathcal{N}(0,1)$.
\end{itemize}
We now contrast the policy gradient update of the abstract action policy, between the 2 options discussed above. 

\paragraph{Raw action space RL.} Performing RL in raw action space, treating the abstract action policy as a model layer, would result in the following expected policy gradient update:

\begin{align}
&    \mathbb{E}\!\left[r_T \sum_{t=1}^T \nabla_\phi \log \pi(a_t \mid s_t, z_t)\right] \nonumber\\
= \;& \mathbb{E}\!\left[r_T \sum_{m=1}^M  \sum_{t=t_m}^{t_{m+1}-1} \nabla_\phi \log \pi(a_t \mid s_t, z_{t_m})\right] \nonumber\\
= \;& \mathbb{E}\!\left[r_T \sum_{m=1}^M  \big[\sum_{t=t_m}^{t_{m+1}-1} \nabla_{z_{t_m}} \log \pi(a_t \mid s_t, z_{t_m}) \big]\nabla_\phi (\mu(s_{t_m} + \epsilon_{t_m})) \right] \nonumber\\
= \;& \mathbb{E}\!\left[r_T \sum_{m=1}^M  \big[\sum_{t=t_m}^{t_{m+1}-1} \nabla_{z_{t_m}} \log \pi(a_t \mid s_t, z_{t_m}) \big]\nabla_\phi \mu(s_{t_m}) \right].\nonumber
\end{align}

\paragraph{Internal RL.} In contrast, internal RL (i.e., RL directly in $z$-space) results in the following policy gradient update:
\begin{align}
& \mathbb{E} [r_T \sum_{m=1}^M \nabla_\phi \log P(z_{t_m} \mid s_{t_m})] \nonumber\\
=\;& \mathbb{E} [r_T \sum_{m=1}^M  \nabla_\phi \frac{-1}{2}(z_{t_m}-\mu(s_{t_m}))^2]\nonumber\\
=\;& \mathbb{E} [r_T \sum_{m=1}^M  \epsilon_{t_m} \nabla_\phi \mu(s_{t_m})].\nonumber
\end{align}
Note that since the forward pass is the same for both methods, the distribution over trajectory and rewards are identical. Let us then compare the variance of these 2 estimators. For simplicity, we now further assume $M=1$. This means a single $z$ is drawn at the beginning of the sequence. The law of total variance gives
\begin{align}
    &\mathbb{V} \big[ r_T \big[\sum_{t} \nabla_{z} \log \pi(a_t \mid s_t, z_0) \big]\nabla_\phi \mu(s_0) \big] \nonumber\\
    =\;&\mathbb{V} \big( \mathbb{E} \big[ r_T  \sum_{t}  \nabla_{z} \log \pi(a_t \mid s_t, z_0) \mid z_0\big]\nabla_\phi \mu(s_0) \big)\nonumber\\
    &+ \mathbb{E}\big( \mathbb{V} \big[ r_T   \sum_{t}  \nabla_{z} \log \pi(a_t \mid s_t, z_0) \mid z_0\big] \nabla_\phi \mu(s_0)\big)\nonumber\\
    =\;&\mathbb{V} \big( \mathbb{E} \big[\text{PG}_{\text{raw}}(z_0)\big] \nabla_\phi \mu(s_0) \big)  + \mathbb{E}\big( \mathbb{V} \big[ \text{PG}_{\text{raw}}(z_0) \big] \nabla_\phi \mu(s_0)\big) \nonumber
\end{align}
where $\text{PG}_{\text{raw}}(z)= r_T  \sum_{t}  \nabla_{z} \log \pi(a_t \mid s_t, z)$ is the raw action space policy gradient Monte Carlo estimator w.r.t.~``parameter'' $z$.

Similarly,
\begin{align}
&\mathbb{V} \big [r_T \epsilon_0 \nabla_\phi \mu(s_0) \big ]\nonumber\\
=\;& \mathbb{V} \big ( \mathbb{E}\big[r_T \mid \epsilon_0 \big] \epsilon_0 \nabla_\phi \mu(s_0) \big ) + \mathbb{E}  \big ( \mathbb{V}\big[r_T \mid \epsilon_0 \big]\epsilon_0 \nabla_\phi \mu(s_0) \big )\nonumber\\
    =\;&\mathbb{V} \big( \mathbb{E} \big[\text{PG}_{\text{z}}(z_0)\big] \nabla_\phi \mu(s_0) \big)  + \mathbb{E}\big( \mathbb{V} \big[ \text{PG}_{\text{z}}(z_0) \big] \nabla_\phi \mu(s_0)\big) \nonumber
\end{align}
where $\text{PG}_{\text{z}}(z)= r_T \epsilon_0$ is the policy gradient Monte Carlo estimator of a bandit problem.

We see that the 2 expressions differ only in $\text{PG}(z)$. The tradeoffs are evident:
\begin{itemize}
\item The expectation of $\text{PG}_{\text{raw}}$ is more structured than $\text{PG}_{\text{z}}$. In particular, its variance w.r.t.~epsilon could even be 0 in the first, whereas it scales with the dimension of epsilon in the second.
\item However, the variance of $\text{PG}_{\text{raw}}$ scales with the number of timestep and with the raw action space dimension, since noise is accumulated at every timestep. On the other hand, $\text{PG}_{\text{z}}$ does not scale with anything (it is the variance of the return, i.e., $O(1)$). Therefore, if the abstract action discovery was successful such that a compact space of $z$ was identified, with long-horizon abstract actions, the policy gradient estimator's variance and corresponding credit assignment can be dramatically improved, especially for very long horizon tasks.
\end{itemize}

\end{document}